\documentclass[acmsmall, hyperref={colorlinks = true,linkcolor=blue, anchorcolor=blue, citecolor=blue}]{acmart}
\usepackage{ulem}
\usepackage{amsmath,bm}
\usepackage{makecell}
\usepackage{subfigure}
\usepackage{multirow}
\usepackage{longtable}
\usepackage{makecell}
\usepackage{url}

\DeclareMathOperator*{\argmin}{arg\,min}

\AtBeginDocument{%
  \providecommand\BibTeX{{%
    \normalfont B\kern-0.5em{\scshape i\kern-0.25em b}\kern-0.8em\TeX}}}

\setcopyright{acmcopyright}
\copyrightyear{2022}
\acmYear{2022}
\acmDOI{XXXXXXX.XXXXXXX}

\acmJournal{JACM}
\acmVolume{37}
\acmNumber{4}
\acmArticle{111}
\acmMonth{8}

\begin{document}

\title{Guided Depth Map Super-resolution: A Survey}

\author{Zhiwei Zhong}

\affiliation{%
  \institution{School of Computer Science and Technology, Harbin Institute of Technology}
  \city{Harbin}
  \country{China}}
\email{zhwzhong@hit.edu.cn}

\author{Xianming Liu}
\authornote{Corresponding Author.}
\affiliation{%
  \institution{School of Computer Science and Technology, Harbin Institute of Technology}
  \city{Harbin}
  \country{China}}
\email{csxm@hit.edu.cn}

\author{Junjun Jiang}
\affiliation{%
  \institution{School of Computer Science and Technology, Harbin Institute of Technology}
  \city{Harbin}
  \country{China}}
\email{jiangjunjun@hit.edu.cn}

\author{Debin Zhao}
\affiliation{%
  \institution{School of Computer Science and Technology, Harbin Institute of Technology}
  \city{Harbin}
  \country{China}}
\email{dbzhao@hit.edu.cn}

\author{Xiangyang Ji}
\affiliation{%
  \institution{Department of Automation and BNRist, Tsinghua University}
  \city{Beijing}
  \country{China}}
\email{xyji@tsinghua.edu.cn}

\begin{abstract}
  Guided depth map super-resolution (GDSR), which aims to reconstruct a high-resolution (HR) depth map from a low-resolution (LR) observation with the help of a paired HR color image, is a longstanding and fundamental problem, it has attracted considerable attention from computer vision and image processing communities. A myriad of novel and effective approaches have been proposed recently, especially with powerful deep learning techniques. This survey is an effort to present a comprehensive survey of recent progress in GDSR. We start by summarizing the problem of GDSR and explaining why it is challenging. Next, we introduce some commonly used datasets and image quality assessment methods. In addition, we roughly classify existing GDSR methods into three categories, i.e., filtering-based methods, prior-based methods, and learning-based methods. In each category, we introduce the general description of the published algorithms and design principles, summarize the representative methods, and discuss their highlights and limitations. Moreover, the depth related applications are introduced. Furthermore, we conduct experiments to evaluate the performance of some representative methods based on unified experimental configurations, so as to offer a systematic and fair performance evaluation to readers. Finally, we conclude this survey with possible directions and open problems for further research. All the related materials can be found at \url{https://github.com/zhwzhong/Guided-Depth-Map-Super-resolution-A-Survey}.
\end{abstract}





\maketitle

\section{Introduction}\label{sec1}

Depth information plays a fundamental role in a wide range of computer vision applications, such as 3D reconstruction~\cite{han2019image, peng2021shape}, semantic scene understanding~\cite{wang2021salient, xu2022weakly, wang2022learning} and autonomous driving~\cite{du2021ago, wang2021progressive}. Currently, there are two main range sensing technologies to capture depth information, either passive or active. For passive sensing technologies, the most famous and commonly used approach is stereo reconstruction, which first obtains two images of the scene meanwhile through two conventional monochrome or color cameras separated by a certain distance and then estimates the scene depth by using stereo matching algorithms. Over the past decades, the performance of stereo matching algorithms has improved markedly~\cite{wang2021pvstereo, shankar2022learned}. However, as these methods are based on finding corresponds between the two views,  the effect of depth missing and occlusion will occur for texture-less regions. In contrast to passive ones, active methods can capture depth maps more efficiently and robustly, especially for texture-less regions. Time-of-flight (TOF) camera~\cite{lange2001solid} and structure light scanner~\cite{herrera2012joint} are two representative depth sensors for real-time active depth sensing. Despite the rapid development of active depth sensing, there is still a large gap between depth camera and color camera, especially in terms of spatial resolution. For instance, the spatial resolution of consumer-level ToF cameras is only about 200 kilo pixels, which is much lower than off-the-shelf conventional color cameras (e.g., 40 Megapixels for Huawei Mate 30). 

The low-resolution depth maps acquired by these depth sensors would greatly affect the performance of algorithms in downstream applications. Therefore, it is highly desirable to develop effective and efficient depth super-resolution algorithms to facilitate the usage of depth information. In general, depth cameras are equipped with an additional RGB sensor which can capture color images with a resolution higher than that of depth maps. It is natural to improve the quality of the depth map by employing the high-resolution color counterpart as guidance, which is known as guided depth map super-resolution (GDSR). %, as shown in Fig.~\ref{fig:fig_1}. 
Compared to the general image super-resolution, the guided depth image super-resolution has two unique characteristics : 1) the depth map contains smooth areas separated by sharp boundaries, which is known as piece-wise smooth, 2) an additional high-resolution color image can be used to guide the depth map reconstruction process. Generally, the depth map and its corresponding color image are geometric and photometric descriptions of the same scene, and they have strong structural similarity. Thus, most existing GDSR algorithms are elaborately designed to exploit this piece-wise smoothness and take full advantage of the guidance image. Recently, GDSR attracts more and more attention in both academic and industrial circles, which encourages us to offer this comprehensive survey of recent advances.
% \begin{figure}[!tb]
%     \centering
%     \includegraphics[width=\linewidth]{figure/f1.pdf}
%     \caption{The flow chart of conventional guided depth map super-resolution (GDSR) algorithms. (b): Visual comparison of the low-resolution depth map and the super-resolved depth map.The purpose of these algorithms is to upsample the input low-resolution (LR) depth map with the guidance of its corresponding high-resolution (HR) color image.}
%     \label{fig:fig_1}
% \end{figure}

\begin{figure*}
    \centering
    \includegraphics[width=\linewidth]{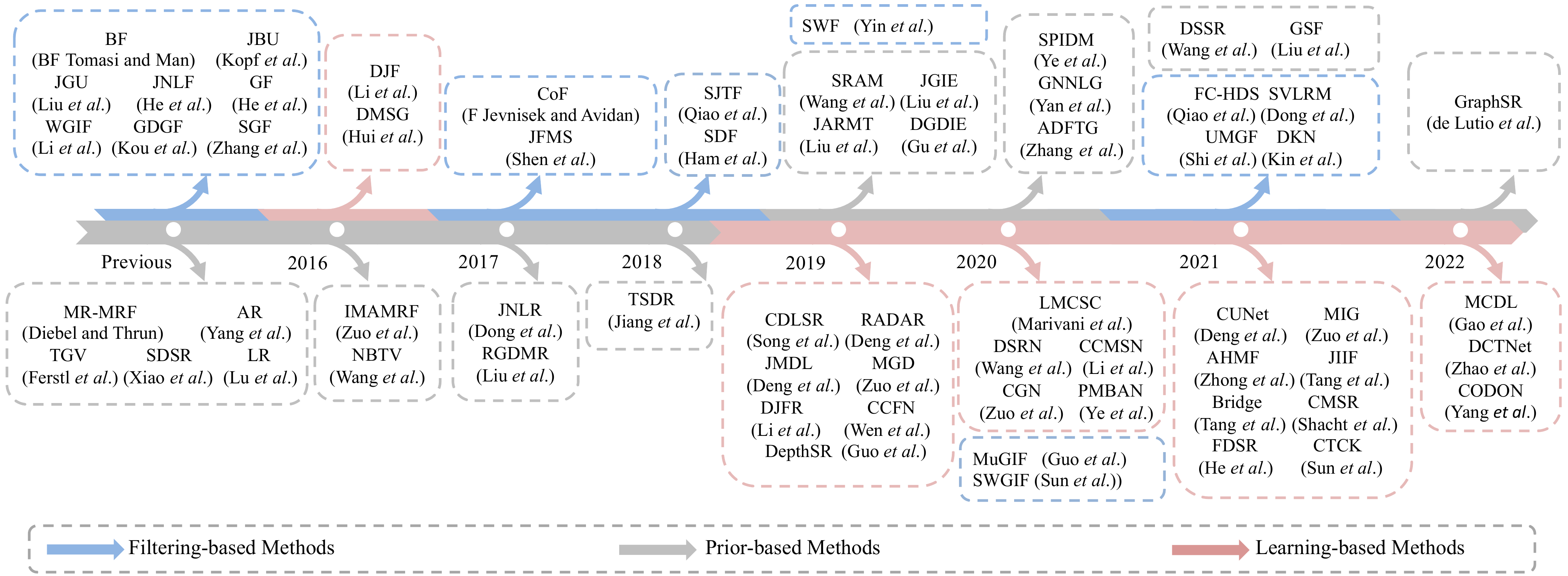}
    \vspace{-0.7cm}
    \caption{Milestones of guided depth map super-resolution methods, including filter-based methods, prior-based methods and learning-based methods.}
    \label{fig:fig2}
    \vspace{-0.2in}
\end{figure*}

Fundamentally, GDSR is an ill-posed problem since there can be various HR depth maps with a slight difference in camera angle, illumination, material properties, and other variables for a certain LR depth map. During the past decades, a wide variety of methods have been developed to solve this inverse problem. To clearly describe the progress of the GDSR methods, we provide a brief chronology in Fig.~\ref{fig:fig2}.
Here, we classify these methods into the following three categories: filtering-based methods, optimization-based methods, and learning-based methods.

Filtering-based methods estimate the depth of pixels by performing a weighted average of local pixels, and the weights are obtained by the affinity calculated from RGB-D image pairs. The bilateral filters~\cite{BF, yang2007spatial, jbu, qiao2021fast}, non-local mean filters~\cite{huhle2010fusion, buades2005non}, and guided filters~\cite{GF, DGF, WGFS, WGF} are representative methods. Filtering-based methods enjoy simplicity in design and low computational complexity, but are prone to generate annoying artifacts when the depth discontinuities are not consistent with those in their corresponding color image. Moreover, the filter kernel is typically designed for a specific task, which lacks flexibility. Contrary to local filtering-based methods, optimization-based methods formulate the problem of GDSR as a global optimization framework. The objective function usually contains two terms: data fidelity term and regularization term. The data fidelity term is used to preserve depth consistency between the super-resolved result and the input depth map. The regularization term encourages the edges of the super-resolved result to consist with the guidance image as much as possible. 
%The major distinction among different approaches in this category is their regularization term, such as Markov random field~\cite{diebel2005application, huhle2007integrating}, auto-regressive model~\cite{yang2012depth, liu2016robust}, total variation~\cite{jiang2018depth} and graph Laplacian~\cite{zhang2019color, yan2020depth, 8491336}. 
The shortcomings of local filtering-based methods can be resolved to some extent by optimization-based methods. Nevertheless, since the optimization-based methods depend on hand-crafted object functions, they may not reflect the real complex image priors. Furthermore, the iterative optimization process involved in these algorithms is usually time-consuming, which impedes their applications in real-world scenarios. To obtain a better and more efficient GDSR model, learning-based methods are proposed and applied to large-scale datasets to solve the depth map super-resolution task. The learning-based methods can be further divided into two categories: sparse dictionary learning methods and deep learning methods. Dictionary learning, also known as sparse coding, aims at deriving a set of dictionary atoms in which only a small number of atoms can be linearly combined to approximate a given image. Due to its simplicity and flexibility for data representation, numerous methods have been introduced in the past decade and achieved impressive results~\cite{9681224, deng2019deep, CUNet}. Recent advances of deep learning algorithms have revolutionized the area of computer vision; tremendous progress has been achieved in a variety of domains, such as image classification~\cite{he2016deep, dosovitskiy2021an}, object localization~\cite{zhang2021weakly, wu2022edn} and semantic segmentation~\cite{strudel2021segmenter, zhao2018psanet}. GDSR is no exception: a myriad of neural networks have been developed and advanced the state of the art~\cite{AHMF, zhao2021discrete, GraphSR}. The learning-based method is suitable for a scene where there is a large amount of training data. When training data are insufficient, the traditional optimization-based or filter-based method may be a better choice.
%In addition, there are some methods~\cite{gu2017learning, gu2019learned, marivani2020multimodal} unfold the steps of traditional optimization algorithms as neural networks to improve the interpretability of the deep model.}

Although significant progress has been made, there is no unified framework to understand and classify existing methods, in particular the recently popular deep learning based approaches. To the best of our knowledge, there are only two surveys~\cite{eichhardt2017image, yang2019depth} available in the literature for the area of GDSR. The paper of \cite{eichhardt2017image} is published in 2017 and its main purpose is to summarize the methods that are tailored for ToF depth maps. The paper of~\cite{yang2019depth} is published in 2019, which only introduces some representative methods that are published before 2019. Many advanced methods that achieve state-of-the-art performance have been proposed since then. Furthermore, neither of these two surveys introduces the benchmark datasets, evaluation metrics, or experimental comparisons of different methods. A comprehensive survey is desired that summarizes and compares these approaches with a unified perspective. In addition, inconsistent experimental protocols are utilized in different approaches. It lacks a benchmark for fairly evaluating state-of-the-art approaches. 
In view of these issues, we provide a comprehensive review of existing GDSR methods. 

We summarize the key contributions of this survey as follows:
\begin{itemize}
    \item We present a systematic investigation of recent GDSR techniques, including problem formulation, benchmark dataset, widely used evaluation metrics, and representative methods. To introduce these methods clearly, we classify them into three categories: filtering-based, optimization-based and learning-based. For each category, we review representative methods and discuss their contributions, benefits, and weaknesses.
    \item We provide comprehensive and fair experimental comparison to evaluate state-of-the-art methods quantitatively and qualitatively on benchmark datasets based on the same experimental configurations. Moreover, we present some tricks that can be used to further improve the performance of depth map reconstruction.
    \item We discuss the open problems, challenges, and envision prospects for future research.
\end{itemize}
% Although there are good reviews of GDSR algorithms~\cite{eichhardt2017image, yang2019depth}, the most recent~\cite{yang2019depth} was published in 2019, and many representative approaches that achieve state-of-the-art performance have been proposed since then; moreover, they paid more attention to traditional optimization-based approaches.

The remainder of this paper is organized as follows. In Sect.~\ref{fund}, we introduce some background concepts and basic knowledge, including the formulation of the GDSR problem, widely used RGB-D datasets, and evaluation metrics. Then detailed reviews of filtering-based (Sect.~\ref{filter}), optimization-based (Sect.~\ref{prior}) and learning-based (Sect.~\ref{learn}) methods are given. In Sect.~\ref{application}, we introduce the depth related applications. In Sect.~\ref{comparison}, we systematically compare several representative methods using publicly available datasets with unified experimental configurations. Finally, we conclude the paper and present some challenging issues as well as promising future research directions in Sect.~\ref{conclusion}.

\section{Fundamentals}\label{fund}
In this section, we introduce the problem formulation of guided depth map super-resolution (GDSR) and commonly used datasets and evaluation metrics. 
\subsection{Problem Formulation}
Current range sensing systems are far from perfect due to a few technical limitations, thus a depth map acquired by these systems is a degraded version of the underlying ground truth depth map. Let $\bm{y}$ and $\bm{x}$ denote the acquired low-resolution (LR) and underlying high-resolution (HR) depth maps, respectively. The observation model in GDSR can be described as follows:
\begin{equation}
    \bm{y} = \bm{H}\bm{x} + \bm{n},
\end{equation}
where $\bm{H}$ denotes the observation matrix and $\bm{n}$ is the introduced additive noise. For different sensing systems, the observation matrix can take different forms. For example, the depth maps captured by the Kinect camera usually experience both structure missing and random depth pixels missing; while the depth maps captured by the ToF camera might suffer both low-resolution and noisy degradations. It is worth noting that in the field of color image processing, a similar formulation has been extensively studied. GDSR differs from them because of the existence of the guidance images. How to take full advantage of the guidance image to improve the performance of GDSR presents new opportunities and challenges to researchers.

To make the mathematical formulations clear, we summarize the notations used in this paper and list them in Table~\ref{tab:symbol}. Here, we use normal-case letters to represent the scalars, upper bold letters to represent the matrices, and lower bold letters to represent the vectors. 
\begin{table}[!t]\setlength{\tabcolsep}{2.5pt}\renewcommand{\arraystretch}{1.2}%\renewcommand\theadfont{\tiny}
    \centering
    % \tiny
     \caption{\footnotesize Summary of commonly used notations in this paper.}
     \vspace{-0.15in}
    \begin{tabular}{lc|lc}
    \toprule
     Notation  & Description &Notation & Description \\
     \hline
    $\bm{X} \in \mathbb{R}^{H\times W}$ & the GT depth map &  $\hat{\bm{Y}} \in \mathbb{R}^{H\times W}$ & the interpolated LR depth map \\
    $\bm{Y} \in \mathbb{R}^{\frac{H}{s}\times \frac{W}{s}}$ & the LR depth map & $\bm{Y}^* \in \mathbb{R}^{H\times W}$ & the super-resolved depth map\\
    $\bm{G} \in \mathbb{R}^{H\times W}$ & the guidance image & $\bm{x, y, g}$ & the vectorized form of $\bm{X, Y}$ and $\bm{G}$, respectively   \\
    \bottomrule
    \end{tabular}
    \label{tab:symbol}
    \vspace{-0.15in}
\end{table}
\subsection{Datasets for GDSR}
\begin{table}[!t]\setlength{\tabcolsep}{4.0pt}\renewcommand{\arraystretch}{1.2}
    \small
    \begin{center}
    \caption{The summary of widely used RGB-D datasets for guided depth map super-resolution.}
    \vspace{-0.15in}
    \label{tab_dataset}
    \begin{tabular}{lccccc}
\toprule
\midrule
% Middlebury 2003~\cite{2003} & 2003& Indoor & - & Stereo Camera & 2 \\
% Middlebury 2005~\cite{2005, 2006} & 2006 & Indoor & - & Stereo Camera & 9 \\
% Middlebury 2006~\cite{2005, 2006} & 2006 & Indoor & - & Stereo Camera & 21\\
Middlebury~\cite{2003, 2005, 2006} & - & Indoor & - & Stereo Camera & 32 \\
ToFmark~\cite{ferstl2013image}& 2013 & Indoor &Time of Flight & PMD Nano & 3 \\
NYU v2~\cite{NYU} & 2012 & Indoor & Structure Light  & Kinect V1& 1449 \\
MPI Sintel Depth~\cite{butler2012naturalistic} & 2012 & Indoor, Outdoor & Animated Film & Blender & 1064\\
Lu~\cite{Lu}& 2014 & Indoor & Structure Light & ASUS Xtion Pro & 6   \\
SUN RGB-D~\cite{SUN} & 2015 & Indoor & Time of Flight & Kinect v2 & 2860 \\
DIDOE~\cite{DIDOE} & 2019 & Indoor, Outdoor & Laser scanner & FARO Focus S350 & 27858\\
DIML Indoor~\cite{cho2021deep} & 2021 & Indoor& Time of Flight & Kinect v2 & 2000   \\
DIML Outdoor~\cite{cho2021deep} & 2021 & Outdoor & Stereo Matching & ZED Stereo & 2000 \\
RGB-D-D~\cite{FDSR} & 2021 & Indoor, Outdoor& Time of Flight & Huawei P30 Pro, Helios & 4811 \\
\bottomrule
\end{tabular}
\end{center}
\vspace{-0.2in}
\end{table}

With the rapid development of GDSR algorithms, a great number of RGB-D datasets have been constructed in the last few decades. We describe 10 RGB-D datasets that are widely used in the literature in Table~\ref{tab_dataset}. The detailed descriptions of each dataset are as follows:

\textbf{Middlebury Dataset}~\cite{2003, 2005, 2006}: Middlebury is a dataset of indoor scenes, which is the most widely used dataset before the deep learning technique is introduced to the field of GDSR. It consists of three small-scale datasets: 2003~\cite{2003}, 2005~\cite{2005, 2006} and 2006~\cite{2005, 2006}. As these datasets do not provide depth maps, previous methods adopted disparity data as depth maps for evaluation. 
% The ground-truth disparities of the 2003, 2005 and 2006 datasets are acquired by a novel technique provided in~\cite{2003} which uses structured lighting that does not require the calibration of the light projectors. To improve the resolution and realism of the existing datasets, ~\citet{2014} propose a 2D subpixel correspondence search algorithm to improve the efficiency of searching and bundle adjustment to improve the accuracy of calibration and rectification. Based on these two novel techniques, the authors propose the 2014~\cite{2014} dataset with 33 high-resolution image pairs. 

\textbf{ToFMark Dataset}~\cite{ferstl2013image}: This dataset consists of three RGB-D image pairs (\textit{i.e.}, \textit{Books, Shark, Devil}) which are captured from real-world scenes. The LR depth maps are captured by a PMD Nano ToF camera, while the HR depth maps are first acquired by using a structured light scanner which is equipped with a high-speed projector. The final ground truth depth maps are then obtained by combining multiple acquisitions with slightly displaced projection angles.

\textbf{NYU v2 Dataset}~\cite{NYU}: NYU v2 is an indoor dataset collected by a Microsoft Kniect v1 sensor. 
%It is the most cited dataset for deep learning based GDSR methods, as its data scale is relatively large than previous datasets. 
This dataset is a continuation of NYU v1~\cite{nyuv1}, which utilizes the same sensor and data structure, but contains fewer scenes and image pairs. 

\textbf{MPI Sintel Depth Dataset}~\cite{butler2012naturalistic}: This dataset is generated from the open source 3D animated short film \textit{Sintel}. MPI Sintel Depth contains more complex and realistic scenes, such as fog, large motions, and blur. 
%The depth values are obtained from Blender in an additional Z-buffer pass. 
It includes 1064 image pairs used for the evaluation of the GDSR algorithm. 

\textbf{Lu Dataset}~\cite{Lu}: It contains six representative RGB-D image pairs which are obtained by the ASUS Xtion Pro camera, and several GDSR methods use it as the testset.

\textbf{SUN-RGB-D Dataset}~\cite{SUN}: This dataset is widely used for scene understanding tasks. It contains 10355 RGB-D image pairs for training and 2860 RGB-D image pairs for testing. The test RGB-D pairs are obtained by Microsoft Kinect v2.% and some works employ them to evaluate the GDSR models.

\textbf{DIDOE Dataset}~\cite{DIDOE}: It is the first public dataset that captures both indoor and outdoor RGB-D image pairs with the same camera. It contains high-resolution RGB images with precise and far-range depth measurements. %In this paper, we use the 753 RGB-D image pairs from the indoor testset to benchmark the GDSR algorithms.

\textbf{DIML Dataset}~\cite{cho2021deep, kim2017deep, kim2018deep, kim2016structure}: DIML is a newly collected large-scale RGB-D dataset which consists of more than 200 indoor / outdoor scenes. For indoor scenes, the authors capture the depth maps using the Kinect v2 sensor. For outdoor scenes, they first use an accurate stereo matching method to generate disparity maps and then convert them to depth maps using the calibration parameters.

\textbf{RGB-D-D Dataset}~\cite{FDSR}: This dataset consists of 4811 image pairs captured from both indoor and outdoor scenes. The "D-D" represents the paired LR and HR depth maps acquired by the Huawei P30 Pro and the Lucid Helios camera.
%To keep the same magnitude of depth map values, the authors choose depth sensors which are all based on the Time-of-Flight sensing technique. %In ACM ICMR 2021, the authors held a real-world depth map super-resolution challenge~\footnote{https://icmr21-realdsr-challenge.github.io} by using this dataset. 

\subsection{Assessment Metrics for GDSR}

For GDSR, quantitative comparison is an essential part of evaluating the performance of different methods. Two of the most frequently used metrics are Mean Absolute Error (MAE), Mean Square Error (MSE) and Root Mean Square Error (RMSE), which are defined as follows:
\begin{equation}
     \text{MAE} = \frac{1}{N} \sum_{i=1}^N \lvert \bm{X}_{i}-\bm{Y}^{*}_i \rvert, \quad \text{MSE} = \frac{1}{N} \sum_{i=1}^N (\bm{X}_{i}-\bm{Y}^{*}_i)^2, \quad \text{RMSE} = \sqrt{\text{MSE}},
\end{equation}
where $\bm{X}_{i}$ means the $i-$th pixel-value of the ground truth depth map, $\bm{Y}^{*}_i$ represents the $i-$th pixel-value of the super-resolved depth map, and $N$ denotes the total number of pixels. Some minority measurements, such as the Structural Similarity Index Measure (SSIM) and Peak Signal-to-Noise Ratio (PSNR) are also sometimes utilized. The PSNR can be calculated as follows:
\begin{equation}
    \text{PSNR} = 10 \cdot \log_{10} (\text{MAX}^2/\text{MSE}),
    \label{ssim}
\end{equation}
where MAX is the maximum pixel value of the depth map. SSIM~\cite{wang2004image} is a traditional indicator of image quality assessment that employs a combination of structures, luminance, and contrast to calculate similarity:
\begin{equation}
    \text{SSIM} = \frac{\left(2 \mu_{x} \mu_{y}+c_{1}\right)\left(2 \sigma_{x y}+c_{2}\right)}{\left(\mu_{x}^{2}+\mu_{y}^{2}+c_{1}\right)\left(\sigma_{x}^{2}+\sigma_{y}^{2}+c_{2}\right)},
\end{equation}
where $\mu_x$ and $\mu_y$ represent the means of the ground truth and the super-revolved depth maps, respectively;  $\sigma_x^2$ and $\sigma_y^2$ denote the variance of the ground truth and the super-revolved depth maps, respectively; $c_1$ and $c_2$ denote two variables which are used to stabilize the division.

\section{Filtering-based Methods}\label{filter}
\begin{table}[!t]\setlength{\tabcolsep}{2.5pt}\renewcommand{\arraystretch}{0.8}\renewcommand\theadfont{\tiny}
    \tiny 
    \begin{center}
    \vspace{-0.04in}
    \caption{A brief summary of the filtering-based methods.}
    \vspace{-0.15in}
    \label{tab_filter}
    \begin{tabular}{cccc}
\toprule
Methods & Published & Category & Basic Idea \\
\midrule
\multirow{10}*{Bilateral filter} & BF~\cite{BF} & ICCV-1998  & \thead[l]{Constructs the kernels by considering both the spatial distance and the variation of intensities.} \\
\cline{2-4}
& JBU~\cite{jbu} & TOG-2007   & \thead[l]{Employs the spatial filter on the LR depth map and the range filter on the HR intensity guidance image.} \\
\cline{2-4}
&JGU~\cite{JGU} & CVPR-2013  & \thead[l]{Uses the geodesic distance instead of the Euclidean distance to quantify the dissimilarities of the pixels.} \\
\cline{2-4}
& CoF~\cite{COF} & CVPR-2017  & \thead[l]{Replaces the range filter in the bilateral filter with a co-occurrence matrix to distinguish the boundaries.} \\
\cline{2-4}
 & JTF~\cite{JTF} & TCYB-2016  & \thead[l]{Constructs the kernels with the spatial distance, color difference and local depth gradient.}\\
%  \cline{2-4}
%  & SJTF~\cite{qiao2018novel} & TMM-2018  & \thead[l]{Constructs the kernels with the spatial distance, color difference and color image segmentation result.} \\
%  \cline{2-4}
% & FC-HDS~\cite{FC-HDS} & MM-2021  &  \thead[l]{Constructs the kernels with the spatial distance, color difference and the classification result of the \\non-blank pixels.}  \\
\hline
 Non-Local Means& JNLF~\cite{huhle2010fusion} & CVPR-2010 & \thead[l]{Follows a similar idea of JBU~\cite{jbu} but calculates the affinity by using image patches instead of single pixels.}\\
\hline
\multirow{13}*{Guided filter} & GF~\cite{GF} & ECCV-2010  & \thead[l]{Assumes that the filtered result is a local linear transform of the guidance image, and the result is obtained\\ by solving a least-square problem.}\\
\cline{2-4}
& WGIF~\cite{WGF} & TIP-2014  & \thead[l]{Integrates an edge-aware weighting into the GF~\cite{GF} to deal with the problem of halo artifacts. }\\
\cline{2-4}
& GDGF~\cite{GDGF} & TIP-2015  & \thead[l]{Introduces an explicit first-order edge-aware constrains into the GF~\cite{GF} to better preserve the edges.} \\
\cline{2-4}
& SKWGIF~\cite{WGFS} & TIP-2020  & \thead[l]{Employs the steering kernel to learn the boundary information and incorporates it into the filter process.}\\ 
\cline{2-4}
& SVLRM~\cite{SVLRM} &  TPAMI-2021  & \thead[l]{Learns the linear representation coefficients ($a_k, b_k$) in the GF~\cite{GF} by a deep convolution neural network.}\\
\cline{2-4}
& UMGF~\cite{UMGF} & TIP-2021  & \thead[l]{A new formulation of guided filter inspired by unsharp masking, and it only needs to estimate one coefficient.} \\
\hline
\multirow{15}*{Dynamic filter} & SGF~\cite{zhang2015segment} & CVPR-2015  & \thead[l]{Segments graph based filter that takes the tree distance calculated on the segment graph as guidance affinity.} \\
\cline{2-4}
& SWF~\cite{SWF} & CVPR-2019  & \thead[l]{A general filtering framework that the neighbors of each pixel can be adjusted dynamically.}\\
\cline{2-4}
& JFMS~\cite{JFMS}& IJCV-2017  & \thead[l]{Introduces the concept of mutual-structure, the goal of image filter can be achieved by correctly transferring\\ the mutual-structures.} \\ 
\cline{2-4}
& SDF~\cite{SDF}& TPAMI-2018  &\thead[l]{Incorporates both static (guidance image) and dynamic (regularized target image) guidance into a unified\\ framework to alleviate the effect of inconsistent structures.} \\ 
\cline{2-4}
& muGIF~\cite{muGIF}& TPAMI-2020 & \thead[l]{Proposes a new measurement to manage the structure similarity of two inputs at the pixel-level and a global\\ optimization objective.}\\
\cline{2-4}
& DKN~\cite{DKN} & IJCV-2021  & \thead[l]{Joint learns the kernel weight and the neighbors by a neural network.} \\
\bottomrule
\end{tabular}
\end{center}
\vspace{-0.2in}
\end{table}
In this section, we introduce the filtering-based methods. Based on the filter kernel construction process, we group existing methods into four categories: bilateral filter~\cite{BF} and its variants, non-local means filter~\cite{buades2005non} and its variants, guided filter~\cite{GF} and its variants, and dynamic filters. The dynamic means that some factors in the filter are dynamic, such as dynamic guidance~\cite{SDF}, dynamic neighbors~\cite{SWF}, etc. Note that for those methods that employ the optimization algorithm or deep convolution neural network for filter kernel generation, we classify them in this category, as they follow the basic framework of the image filters. A brief summary of the filtering-based methods can be found in Tab.~\ref{tab_filter}.
\subsection{Bilateral Filters}
The bilateral filter (BF), originally introduced by~\cite{BF} is a kind of image filter that aims to smooth the image as well as preserve the edge. The basic idea of BF is to take into account both the spatial distance and the variation of intensities when constructing the filter kernels. Formally, for a pixel at the index $p$, the filtered output can be obtained by the following equation:
\begin{equation}
    \bm{Y}^*_p = \frac{1}{\text{W}_p} \sum_{q \in \mathcal{N}_p} f(\|p-q\|_2^2) g(\|\bm{Y}_p- \bm{Y}_q\|_2^2) \bm{Y}_{q},
\end{equation}
where $\bm{Y}_p$ and $\bm{Y}^*_p$ are the input and output images at the pixel index $p$; $\mathcal{N}_p$ are the neighbors of the pixel $p$; $f(\cdot)$ and $g(\cdot)$ are two Gaussian kernels;
%which are defined as follows:
% \begin{equation}
%     f(\|p-q\|_2^2) = \exp \left(\frac{-\|p-q\|_{2}^{2}}{2 \sigma_{1}^{2}}\right), \quad g(\|\bm{Y}_p- \bm{Y}_q\|_2^2)= \exp \left(\frac{-\|\bm{Y}_p- \bm{Y}_q\|_{2}^{2}}{2 \sigma_{2}^{2}}\right)\red{.}
% \end{equation}
% where $\sigma_1$ and $\sigma_2$ are the kernel bandwidth parameters; 
$\mathrm{W}_p = \sum_{q\in \mathcal{N}_p} f(\|p-q\|_2^2) g(\|\bm{Y}_p- \bm{Y}_q\|_2^2)$ denotes the normalization factor. Following \cite{BF}, \citet{JBF} propose a joint bilateral filter (JBF) which applies the range filter the guidance image $\bm{G}$, and it can be formulated as follows:
\begin{equation}
    \bm{Y}^*_p = \frac{1}{\text{W}_p} \sum_{q \in \mathcal{N}_p} f(\|p-q\|_2^2) g(\|\bm{G}_p- \bm{G}_q\|_2^2) \bm{Y}_{q}.
\end{equation}
%To enhance the resolution of depth maps, \citet{4270236} propose a novel post-processing step by applying JBF~\cite{JBF} to the 3D cost volume iteratively. 
Motivated by JBF~\cite{JBF}, ~\citet{jbu} propose a joint bilateral upsampling (JBU) to restore a high-resolution (HR) depth map from a low-resolution (LR) observation with an HR color image as guidance. JBU~\cite{jbu} uses the same kernel as JBF~\cite{JBF}, and it can be expressed as follows:
\begin{equation}
    \bm{Y}^*_p = \frac{1}{\text{W}_p} \sum_{q \in \mathcal{N}_p} f(\|p_\downarrow - q_\downarrow \|_2^2) g(\|\bm{G}_p- \bm{G}_q\|_2^2) \bm{Y}_{q_\downarrow}.
\end{equation}
where $p$ and $q$ represent the coordinates in the HR image; $p_{\downarrow}$ and $q_{\downarrow}$ mean the corresponding coordinates (possibly fractional) in the LR image. The bilateral filter cannot determine whether an edge is within a texture or not, as it is defined by a spatial Gaussian and a range Gaussian filter. \citet{COF} propose co-occurrence filter (CoF) to address this problem. In particular, they propose a normalized co-occurrence matrix that can represent the co-occur frequency of pixel values to replace the range Gaussian filter used in the BF~\cite{BF}, and the CoF~\cite{COF} is given by:
\begin{equation}
    \bm{Y}^*_p = \frac{1}{\text{W}_p} \sum_{q \in \mathcal{N}_p} f(p, q) \frac{\bm{C}(a, b)}{h(a)h(b)} \bm{Y}_{q_\downarrow},
\end{equation}
where $f(\cdot)$ is a Gaussian filter kernel; $h(a)=\sum_p[\bm{I}^\text{g}_a=a]$, $[\cdot]$ equals 0 if the expression in the brackets is false and 1 otherwise; $\bm{C}$ is a co-occurrence matrix, and the co-occurrence value of a and b can be obtained by the following equation:
\begin{equation}
    \bm{C}(a, b) = \sum_{p, q} f(\|p - q\|_2^2)[\bm{G}_a=a][\bm{G}_b=b].
\end{equation}
To simplify the computation, the authors calculate the co-occurrence matrix only in a local window.

Instead of employing the Euclidean distance as the JBU~\cite{jbu}, \citet{JGU} propose a joint geodesic upsampling (JGU) which uses the geodesic distance to measure the affinity between two coordinates. 
% The geodesic distance between p and q is defined as the length of the shortest path:
% \begin{equation}
%     \begin{aligned} d(p, q)=& \min _{k \in \mathcal{K}} \sum_{i=2}^{\lvert k \rvert}\left(\frac{1}{r}\left\|p_{k}^{(i)}-p_{k}^{(i-1)}\right\|^2_{2}\right. \left.+\lambda\left\|\bm{G}_{p_{k}^{(i)}} - \bm{G}_{p_{k}^{(i-1)}}\right\|^2_{2}\right), \end{aligned}
% \end{equation}
% where $k$ means a path that joins $p$ and $q$; $\lvert k \rvert$ represents the number of pixels in path $k$, and $\mathcal{K}$ denotes the set of all the paths that join $p$ and $q$; $r$ is the upsampling factor. 
% Based on the geodesic distance, we can calculate the affinity between $p$ and $q$ as follows:
% \begin{equation}
%     f(\|p-q\|_2^2) = \exp(-d(p, q) / 2\sigma^2),
% \end{equation}
% where $\sigma$ is the kernel bandwidth parameter. The JGU~\cite{JGU} can avoid the depth missing problem to some extent; however, its results still suffer from depth bleeding artifacts.
Existing methods are based on the assumption that the discontinuities in depth and the guidance are consistent. Actually, this assumption may not always be true, and this will lead to the problem of texture copying.  To address this issue, \cite{JTU} propose a joint trilateral upsampling (JTU) that generates filter kernels by incorporating edge information from the LR depth map:
\begin{equation}
    f(\|p_\downarrow - q_\downarrow\|_2^2) g(\|\bm{G}_{p} - \bm{G}_{q}\|_2^2) h(\|\bm{Y}_{p_\downarrow} - \bm{Y}_{q_\downarrow}\|_2^2).
    \label{eq_jtu}
\end{equation}
% Moreover, to preserve sharp boundaries and remove noise, the authors propose an iterative implementation in which the depth map of the previous step is fed into the current step. 
\citet{JTF} propose a similar idea that advances the JBF~\cite{JBF} with the LR depth map for kernel construction, named Joint Trilateral Filter (JTF). The kernel function of JTF~\cite{JTF} can be obtained by replacing the third term of Eq.~\ref{eq_jtu} with the gradient map of the interpolated LR depth map.  JTF~\cite{JTF} can generate sharp edges for lower upsampling factors, but the performance will drop when the upsampling factor becomes large. \citet{qiao2018novel} propose to use the result of the segmentation of the guidance image as one of the weight terms for kernel generation, as they find that the segmented regions can adhere well to the depth boundary. 
% \citet{FC-HDS} propose a fast classification-based hierarchical method for depth map super-resolution (FC-HDS), and to improve the calculation speed, they perform filtering only near the depth edges through matrix operations. Furthermore, they propose to integrate the classification result into the kernel function of JBU~\cite{jbu} to increase the contributions of reliable neighbors. 

\subsection{Non-Local Means Filters}
The basic idea of the non-local means (NL-Means) filter~\cite{buades2005non} is similar to the BF~\cite{BF} method which restores a pixel as a weighted average of its neighbor pixels.
% \begin{equation}
%     \bm{Y}^*_p = \frac{1}{\text{W}_p} \sum_{q \in \mathcal{N}_p} w(\bm{Y}_p, \bm{Y}_q) \bm{Y}_{q}.
%     \label{nl_1}
% \end{equation}
However, the definition of the kernel function $w(\cdot)$ is slightly different. For the NL-Means filter, the similarities are determined by the patches surrounding both pixels instead of the single-pixel values, which is formulated as
\begin{equation}
    w(\bm{Y}_p, \bm{Y}_q) = \exp (-\frac{1}{h} \sum_{k\in N} f (\|k\|) (\bm{Y}_{p+k} - \bm{Y}_{q+k})^2),
\end{equation}
where $h$ means a smoothing parameter; $N$ denotes the number of pixels in a patch and $k$ is the offset of the center pixel. Inspired by \cite{jbu} and the NL-Means filter~\cite{buades2005non}, \citet{huhle2010fusion} propose a joint non-local means filter (JNLF), which has the following form:
\begin{equation}
    \bm{Y}^*_p = \frac{1}{\text{W}_p} \sum_{q \in \mathcal{N}_p} \xi_{pk} w(\bm{G}_p, \bm{G}_q) w(\bm{Y}_{p_\downarrow}, \bm{Y}_{q_\downarrow}) \bm{Y}_{q_\downarrow},
\end{equation}
where $\xi_{pk}=\exp{(-(\bm{Y}_{p_\downarrow} - \bm{Y}_{(p+k)_\downarrow})^2) / h}$ represents a weighting factor to restrict affinity comparison within areas of similar depths.

\subsection{Guided Filters}
The widely used bilateral filter~\cite{BF} and its variants may suffer from gradient reversal artifacts, especially for detail enhancement tasks~\cite{farbman2008edge, durand2002fast, bae2006two}, and are typically time-consuming. To alleviate these problems, \citet{GF} propose a novel translation-variant image filter named guided filter (GF) based on the assumption that the filtered result $\bm{Y}^*$ can be obtained from the guidance image $\bm{G}$ by a local linear transform in a window $\mathcal{N}$, which can be expressed as follows:
\begin{equation}
    \bm{Y}^*_p = a_k \bm{G}_p + b_k, \forall k \in \mathcal{N}_k,
    \label{eq_gf}
\end{equation}
where $a_k$ and $b_k$ are two linear coefficients that are assumed to be constant values in the window $\mathcal{N}_k$ centered on the pixel $k$. The values of these two linear coefficients can be obtained by minimizing an objective function that calculates the difference between the filtered result and the input image:
\begin{equation}
    \argmin_{a_{k}, b_{k}} \sum_{p \in \mathcal{N}_k} \left((a_k\bm{G}_p + b_k)^2 -  \bm{Y}_p + \gamma a_{k}^{2} \right),
    \label{eq_gf2}
\end{equation}
where $\gamma$ denotes a regularization parameter that is utilized to penalize large values for $a_k$.  By solving a least-squares problem of Eq.~\ref{eq_gf2}, we can obtain the optimal values of $a_k$ and $b_k$, and they are:
\begin{equation}
    a_k = \frac{\frac{1}{\lvert \mathcal{N} \rvert} \sum_{p\in \mathcal{N}_k} \bm{G}_p \bm{Y}_p - \overline{\bm{G}}_k \overline{\bm{Y}}_k}{\sigma_k^2 + \gamma}, \quad b_k = \overline{\bm{Y}}_k - a_k \overline{\bm{G}}_k,
    \label{eq_gf_b}
\end{equation}
where $\overline{\bm{G}}_k$ and $\overline{\bm{Y}}_k$ are the mean of the guidance and target images in $\mathcal{N}_k$, respectively; $\sigma_k^2$ denotes the variance of the guidance image in $\mathcal{N}_k$; $\lvert \mathcal{N} \rvert$ means the number of pixels in $\mathcal{N}$. As a pixel $p$ can be involved in different windows, which may produce different values of $\bm{Y}^*_p$ in Eq.~\ref{eq_gf}. The authors propose a naive strategy that averages all possible values of $\bm{Y}^*_p$ as the final filtered result:
\begin{equation}
    \bm{Y}^*_p = \overline{a}_p \bm{G}_p + \overline{b}_p,
    \label{eq_gf_3}
\end{equation}
where $\overline{a}_p=\frac{1}{\lvert \mathcal{N}\rvert} \sum_{k\in \mathcal{N}_p} a_k$ and $\overline{b}_p=\frac{1}{\lvert \mathcal{N}\rvert} \sum_{k\in \mathcal{N}_p} b_k$. 
% Based on this, \citet{DGF} extend the GF\cite{GF} to a deep learning framework, namely the deep guided filtering network (DGF). Specifically, they formulate the guided filter algorithm as a fully differentiable layer. This layer can be seamlessly plugged into a convolution neural network and trained end-to-end. 

As an edge-preserving smoothing filter, the GF~\cite{GF} not only prevents gradient reversal artifacts but also enjoys the merit of high computational efficiency. However, it may suffer from halo artifacts near some edges because of the filter's local character, and the regularization parameter is set as a constant. To address these problems, \citet{WGF} propose a guided image filter (WGIF) which incorporates an edge-aware weighting into the GF~\cite{GF} to dynamically adjust the regularization parameter. The edge-aware weighting $\Gamma\left(k\right)$ is calculated by integrating local and global information from the guidance image, and is defined as follows:
\begin{equation}
    \Gamma\left(k\right) = \frac{1}{N} \sum_{i=1}^N \frac{\sigma^2_k + \varepsilon}{\sigma^2_i + \varepsilon},
\end{equation}
where $\sigma^2_k$ and $\sigma^2_i$ denote the variance of the guidance image in a $3\times 3$ window centered on the pixel $k$ and $i$, respectively; $\varepsilon$ means a small constant, and the authors set it as ($0.001 \times L)^2$ where $L$ represents the dynamic range of the input target image. After obtaining the weight, the authors incorporate it into the object function of the GF~\cite{GF}, which can be expressed as follows:
\begin{equation}
    \argmin_{a_{k}, b_{k}} \sum_{p \in \mathcal{N}_k} \left((a_k \bm{G}_p + b_k)^2 -  \bm{Y}_p + \frac{\gamma}{\Gamma\left(k\right)} a_{k}^{2} \right).
    \label{eq_wgf}
\end{equation}
% The optimal values of $a_k$ and $b_k$ are as follows:
% \begin{equation}
%     a_k = \frac{\frac{1}{\lvert \mathcal{N} \rvert} \sum_{p\in \mathcal{N}_k} \bm{G}_p \bm{Y}_p - \overline{\bm{G}}_k \overline{\bm{Y}}_k}{\sigma_k^2 + \frac{\gamma}{\Gamma\left(k\right)}}, \quad b_k = \overline{\bm{Y}}_k - a_k \overline{\bm{G}}_k.
% \end{equation}
Compared to the GF~\cite{GF}, the WGIF~\cite{WGF} can produce sharper edges and alleviate halo artifacts. However, as mentioned in \cite{WGFS}, the GF~\cite{GF} and the WGIF~\cite{WGF} tend to produce results with blurred boundaries, as they adopt the average strategy to calculate the final result. In this average strategy, the weights of all pixels in a local window are assigned with the same values, and the boundary directions will easily be neglected. To solve this problem and make better use of the boundary information, \citet{WGFS} propose to incorporate the structure priors which are learned from the guidance image by using the steering kernel~\cite{takeda2007kernel} into the WGIF~\cite{WGF}.

In the GF~\cite{GF}, the authors assume that the filtered result is a local linear transformation of the guidance image as written in Eq.~\ref{eq_gf}. According to this, we can see that the gradient of the target and the guidance images in a local window would satisfy the following equation:
\begin{equation}
    \nabla \bm{Y}_p^* = a_k \nabla \bm{G}_p.
    \label{eq_svlrm_1}
\end{equation}
The Eq.~\ref{eq_svlrm_1} guarantees that the target image has the same structure as the guidance image since the coefficient $a_k$ is constant for each local window. Therefore, inconsistent textures in the guidance image will be transferred to the target image, and this would lead to texture-copying artifacts. Furthermore, the GF~\cite{GF} employs the mean filter to obtain the filtered output, which may blur the edges. To address this problem, ~\citet{SVLRM} propose a spatially variant linear representation model (SVLRM) that uses a deep convolution neural to learn the linear representation coefficients. The filtering process can be represented by the following equation:
\begin{equation}
    \bm{Y}_p^* = a(\bm{G}, \bm{Y}) \bm{G} + b (\bm{G}, \bm{Y}),
\end{equation}
where $\bm{I}^\text{g}$ and $\bm{I}^\text{t}$ denote the spatially variant linear representation coefficients which are learned by a deep neural network. The SVLRM~\cite{SVLRM} leverages the neural network to simultaneously estimate two coefficients of the guided filter, this is suboptimal and tends to generate inconsistent edges. The traditional unsharp masking~\cite{morishita1988unsharp} technique can enhance the edge by estimating only one coefficient. Inspired by this, \citet{UMGF} propose a novel and simplified formulation of the GF~\cite{GF}, namely the unsharp mask guided filter (UMGF). The unsharp masking~\cite{morishita1988unsharp} technique can be used to enhance image sharpness, which is described as follows:
\begin{equation}
    \bm{Y}^* = \lambda \left(\bm{Y} - \mathcal{F}_L(\bm{Y})\right) + \bm{Y},
\end{equation}
where $\bm{Y}^*$ denotes the enhanced image and $\bm{Y}$ is the input image; $\mathcal{F}_L$ is a low-pass filter; $\left(\bm{Y} - \mathcal{F}_L(\bm{Y})\right)$ represents the unsharp mask; $\lambda$ is a coefficient to control the degree of enhancement. The filtered result of the GF~\cite{GF} in Eq.\ref{eq_gf_3} can be rewritten as follows:
\begin{equation}
    \bm{Y}^* = \frac{1}{\lvert \mathcal{N} \rvert} \sum_{k\in \mathcal{N}_p} (a_k \overline{\bm{G}}_p +b_k).
    \label{eq_umgf_1}
\end{equation}
To eliminate $b_k$, we can put Eq.~\ref{eq_gf_b} in Eq.~\ref{eq_umgf_1}:
\begin{equation}
    \bm{Y}^* = \frac{1}{\lvert \mathcal{N} \rvert } \sum_{k\in \mathcal{N}_p} a_k \bm{G}_p + \frac{1}{\lvert \mathcal{N} \rvert }\sum_{k\in \mathcal{N}_p} \left(\overline{\bm{Y}}_p - a_k \overline{\bm{G}}_k\right).
\end{equation}
Then, the equation can be changed to the following equation:
\begin{equation}
    \bm{Y}^* = \frac{1}{\lvert \mathcal{N} \rvert } \sum_{k\in \mathcal{N}_p} a_k (\bm{G}_p - \overline{\bm{G}}_k) + \widetilde{\bm{Y}}_p,
    \label{eq_umgf_2}
\end{equation}
where $\widetilde{\bm{Y}}_p =\frac{1}{\lvert \mathcal{N} \rvert}\sum_{k \in \mathcal{N}_p} \bm{Y}_p$. As $\overline{\bm{Y}}$ is the output of a mean filter, we can assume that $\overline{\bm{G}}_p$ is close to its mean in a local window $\mathcal{N}_p$. Thus, Eq.~\ref{eq_umgf_2} can be rewritten as follows:
\begin{equation}
    \bm{Y}^* = \overline{a}_p (\bm{G}_p - \widetilde{\bm{G}}_p) + \overline{\bm{Y}}_p,
    \label{eq_umgf_3}
\end{equation}
where $\overline{a}_p = \frac{1}{\lvert \mathcal{N} \rvert}\sum_{k \in \mathcal{N}_p} a_k$, and $\widetilde{\bm{G}}_p = \frac{1}{\lvert \mathcal{N} \rvert}\sum_{k \in \mathcal{N}_p} \overline{\bm{G}}_k$. From Eq.~\ref{eq_umgf_3}, we can see that, for guided image filtering, the target image is first smoothed to remove redundant information, such as noise and textures. Next, the structure information $(\bm{G}_p - \widetilde{\bm{G}}_p)$ is then transferred from the guidance image to the target image under the control of $\overline{a}_p$. The transferred structures are called the unsharp mask, which was mentioned in the previous part. The Eq.~\ref{eq_umgf_3} can be described in a more general way:
\begin{equation}
    \bm{Y}^* = f_a(\bm{Y}_m, \bm{G}_m) \odot \bm{G}_m + \mathcal{F}_L (\bm{Y}),
\end{equation}
where $\bm{Y}_m=\bm{Y} - \mathcal{F}_L(\bm{Y})$ and $\bm{G}_m=\bm{G} - \mathcal{F}_L(\bm{G})$ are the unsharp masks calculated from the target and the guidance images, respectively; $f_a$ is a function of controlling the degree of structure to be transferred. In this work, the authors propose to learn the function $f_a$ using a deep neural network.
%to boost the reconstruction performance.

\subsection{Dynamic Filters}
Previous image filters are based on the assumption that the structures between guidance image and depth image is consistent. However, inconsistent or even completely different structures usually exist in the two images. Directly adopting this guidance is prone to generate unsatisfactory results. 

To solve this problem, ~\citet{SDF} propose a static and dynamic filter (SDF) to take advantage of static and dynamic guidance. The static guidance shares a similar idea with existing image filters, which modulates the target image with a weight function obtained from the guidance image. The dynamic guidance is derived from the regularized images, i.e., iteratively using the already regularized images as the dynamic guidance to constrain the output with the aim of alleviating the effects of inconsistent structures. 
% The objective function of SDF~\cite{SDF} is as follows:
% \begin{equation}
%     \mathcal{E}(\hat{\bm{Y}^*}) = \sum_p \bm{C}_p (\bm{Y}^*_p - \bm{Y}_p) + \lambda \sum_{p, q\in \Omega} \phi_u(\bm{G}_p - \bm{G}_q)\psi_v(\bm{Y}^*_p - \bm{Y}^*_q),
% \end{equation}
% where the first term is the fidelity term, while the second is the regularization term; $\bm{C}_p$ is a confidence matrix; $\phi_u(x)=\exp(-\mu x^2)$ and $\mu$ is the bandwidth parameter; $\psi_v(x) = (1 - \phi_v(x))/v$. The regularization term consists of both static and dynamic regularizations. 
However, this strategy cannot make full use of the static guidance image and can only handle at the same time. \citet{JFMS} propose the concept of mutual-structure to improve the performance of joint processing in the restoration of common structures. In \cite{JFMS}, the authors define three types of structure: 1) Mutual structure: the common edges contained in the corresponding two patches; 2) Inconsistent structure: the different structures contained in the two patches; 3) Smooth regions: the common low-variance smooth areas in the two patches. Obviously, the goal of the image filter is to transfer the mutual-structures while preventing the inconsistent ones. The mutual-structure of two patches can be measured by:
\begin{equation}
    \mathcal{S}(\bm{Y}^*_p, \bm{Y}_p) = \left(\sigma(\bm{Y}^*_p)^2 + \sigma(\bm{G}_p)^2 \right)\left(1-cov(\bm{Y}^*_p, \bm{G}_p) / \sqrt{\sigma(\bm{Y}^*_p)\sigma(\bm{G}_p)} \right)^2,
\end{equation}
where $\sigma(\bm{Y}^*_p)$ denotes the variance of a patch in a local window centered on the pixel $p$, and $cov(\cdot)$ means the covariance of the patch intensity. The patch-level filtering can be achieved by optimizing this equation. Compared to the SDF~\cite{SDF}, this method can process two inputs simultaneously. However, the results of this method usually suffer from halo artifacts because of its local nature. To solve this problem, \citet{muGIF} propose a mutually guided filtering (muGIF), which designs a global optimization method to improve filtering performance. 
% In addition, they propose a new measurement for structure similarity at the pixel level, called mutual response, which is as follows:
% \begin{equation}
%     \mathcal{R} (\bm{Y}, \bm{G}) = \sum_p \sum_{d \in \{h, v\}} \frac{\lvert\nabla_d \bm{Y}_p\rvert}{\lvert\nabla_d \bm{G}_p\rvert},
% \end{equation}
% where $\nabla_h$ and $\nabla_v$ mean the first-order derivative of the horizontal and vertical directions, respectively. The object function of muGIF~\cite{muGIF} is given by:
% \begin{equation}
%     \argmin_{\bm{Y}^*, \bm{G}^*} \alpha_t \mathcal{R} (\bm{Y}, \bm{G}) + \beta_t\|\bm{Y}^* - \bm{Y}\|^2 + \alpha_g \mathcal{R} (\bm{G}, \bm{Y}) + \beta_g\|\bm{G}^* - \bm{G}\|^2,
%     \label{eq_mugif_1}
% \end{equation}
% where $\alpha$ and $\beta$ are non-negative constants to balance its corresponding terms. Eq.~\ref{eq_mugif_1} can be solved in an alternative manner by decomposing it into several subproblems. 
In addition to the guidance, the neighbors can also be adjusted dynamically. In \cite{SWF}, the authors find that the edge pixels can be well recovered by a weighted average process if its neighboring pixels come from the same side of the edge. To achieve this goal, they propose a side window fitering (SWF) technique where the side window of each pixel can be obtained by solving an optimization problem.
% \begin{equation}
%     \argmin_{\theta, r} \|\bm{Y}_p - \mathcal{F}(\bm{G}_p, \theta, r) \|.
% \end{equation}

\section{Optimization-based Methods}\label{prior}
In this section, we introduce the optimization-based methods. These methods formulate the guided depth map super-resolution (GDSR) as an optimization problem and introduce various priors to make the ill-posed problem traceable. Although the technique details are different, most of the methods in this category share a similar framework, which can be formalized as follows:
\begin{equation}
    \bm{y}^* = \argmin_{\bm{y}^*} \|\bm{y} - \bm{H}\bm{y}^*\| + \lambda \phi(\bm{y}^*),
\end{equation}
where $\bm{y}^* \in \mathbb{R}^N$ is the recovered high-resolution (HR) depth map with $N$ pixels, and $y$ is the observed low-resolution (LR) depth map. $\lambda$ is a weight to balance the data term and the regularization term, and $\phi(\bm{y}^*)$ means the regularization term to smooth the result $\bm{y}^*$ and ensure that it has the same structure as the guidance image $\bm{g}$. The major distinction among the different approaches in this category is their regularization term. We present a brief summary of existing optimization-based methods in Tab.~\ref{tab_prior} and the technique detail of each method can be found in the following part.

\subsection{Markov Random Field}
Markov Random Field (MRF) is an undirected graphical model that can be used to represent a joint probability distribution. It has been extensively applied in many computer vision and image processing tasks, such as semantic segmentation, image restoration, and depth map super-resolution. The MRF can be solved by optimizing the Gibbs energy function on a graph $\mathcal{G}=\langle\mathcal{V}, \mathcal{E}\rangle$~\cite{clifford1990markov}:
\begin{equation}
    E = \sum_{p \in \mathcal{V}}U(\bm{y}_p) + \lambda\sum_{(p, q)\in \mathcal{E}} V(\bm{y}_p, \bm{y}_q),
\end{equation}
where $\mathcal{V}$ means the set of all pixels in an image and $\mathcal{E}$ means the set of edges that connect $p$ and $q$. $U(\bm{y}_p)$ and $V(\bm{y}_p, \bm{y}_q)$ denote the data term and the regularization term, respectively. Based on this, \citet{10.5555/2976248.2976285} propose a multi-resolution MRF (MR-MRF) framework for GDSR. In the MR-MRF, the depth map super-resolution is modeled as a pixel-labeling problem, and the MRF is utilized to estimate the true label for each pixel of the coarse depth map. The object function of MR-MRF~\cite{10.5555/2976248.2976285} is as follows:
\begin{equation}
    E = \sum_{p \in \mathcal{V}}(\bm{y}^*_p - \bm{y}_p)^2 + \lambda\sum_{(p, q)\in \mathcal{E}} (\bm{y}^*_p - \bm{y}^*_q)^2\exp{\left(-c(\bm{g}_p - \bm{g}_q)^2 \right)},
    \label{mr_mrf_1}
\end{equation}
where $\bm{y}^*$ is the super-resolved depth map, $\bm{y}$ is the LR depth map and $\bm{g}$ is the guidance image; $c$ is a constant value that quantifies the degree of edge smoothing. 
%As the size of $\bm{y}$ is not always equal to $\bm{y}^*$ for the super-resolution task, the authors apply %Eq.~\ref{mr_mrf_1} only to a sparse grid of pixels for which the depth measurements are available. 
\citet{lu2011revisit} further extend this work by introducing a new data term that is more suitable for the characteristics of the depth map. The quadratic distance adopted in Eq.~\ref{mr_mrf_1} is prone to generate results with blurred-edge artifacts, to solve this problem, \citet{6637884} propose to use exponential-type functions to define the data term and the regularization term. Previous methods for GDSR are based on the assumption that the edges between the depth map and its corresponding guidance image are consistent. However, this assumption is not always true, and texture-copying artifacts will be introduced when the assumption is invalid. To overcome this problem, \citet{zuo2016explicit} propose a quantitative score to measure the inconsistency between the depth map and its corresponding color image. This score can be integrated into the regularization term of the MRF-based framework to alleviate texture-copying artifacts and preserve depth boundaries. The cost function of \cite{zuo2016explicit} is as follows:
\begin{equation}
    E = \sum_{p \in \mathcal{V}}(\bm{y}^*_p - \bm{y}_p)^2 + \lambda\sum_{(p, q)\in \mathcal{E}} (\bm{y}^*_p - \bm{y}^*_q)^2\exp{\left(-\frac{\lvert \nabla\bm{g}_{pq} \cdot (1 - \alpha_{pq}) \rvert + \lvert \nabla \hat{\bm{y}}_{pq} \cdot \alpha_{pq} \rvert}{2\sigma^2} \right)},
\end{equation}
where $\nabla\bm{g}_{pq}$ and $\nabla\bm{y}^*_{pq}$ denote the difference between the pixel $p$ and $q$ for the guidance image and the coarsely interpolated depth map $\hat{\bm{y}}_{pq}$, respectively; $\sigma$ is a constant value; $\alpha$ is a confidence map that is used to describe the inconsistency between the depth map and its corresponding color image, and the detailed calculation process can refer to the original paper~\cite{zuo2016explicit}. Existing MRF-based GDSR methods that calculate the similarity in the regularization term only based on the differences between the pixel and its neighbors. This strategy neglects the local structure on the depth map and tends to generate the depth map with blurred edges. In order to better maintain the depth edges, \citet{8340850} propose to calculate the similarity in the minimum spanning trees space where the paths in this space can represent the local structure of the depth map. 
\begin{table}[!t]\setlength{\tabcolsep}{2.5pt}\renewcommand{\arraystretch}{0.8}\renewcommand\theadfont{\tiny}
    \tiny 
    \begin{center}
    \vspace{-0.04in}
    \caption{\footnotesize A brief summary of the prior-based methods.}
    \vspace{-0.09in}
    \label{tab_prior}
    \begin{tabular}{lccc}
\toprule
Category & Methods & Published &  Basic Idea \\
\midrule
 \multirow{7}*{Markov Random Field} & MR-MRF~\cite{10.5555/2976248.2976285} & NIPS-2005 & \thead[l]{The pioneering work that utilizes Markov Random Field for multi-modal image fusion.} \\
 \cline{2-4}
~ & IMAMRF~\cite{zuo2016explicit} & TCSVT-2016 &  \thead[l]{Proposes a quantitatively metric to measure the inconsistency between the RGB-D pairs, and embeds\\ this metric into the MRF framework to alleviate texture-copying artifacts.}\\
 \cline{2-4}
~ &  MSFEE~\cite{8340850} & TIP-2018 & \thead[l]{Proposes to compute the guidance affinity by using the tree distance to maintain the depth boundaries.} \\
\hline
\multirow{9}*{Auto-regressive Model} & AR~\cite{6827958} & TIP-2014 &  \thead[l]{Proposes an adaptive color-guided AR model for depth recover where the AR coefficients are derived\\ from the local correction of the LR depth map and non-similarity of the guidance image.} \\
\cline{2-4}
~ & TSDR~\cite{jiang2018depth} & TIP-2018 & \thead[l]{A method which includes both transform and spatial domain regularizers. The transform domain\\ regularizer is a patch-based AR model to characterize intra-pacth correlations, and the spatial domain\\ regularizer is a multi-directional TV to capture the geometrical structures.} \\
\cline{2-4}
~ & SRAM~\cite{wang2019multi} & TMM-2019 &  \thead[l]{A method which combines multi-direction sparse representation and AR to represent depth edges in\\ both patch and pixel level. }\\
\cline{2-4}
\hline
\multirow{6}*{Total Variation} & TGV~\cite{ferstl2013image} & ICCV-2013 & \thead[l]{Employs the second-order TGV as the regularization for GDSR.}\\
\cline{2-4}
% ~ & SDSR~\cite{xiao2015defocus} & CVPR-2015 & \thead[l]{Simultaneously \red{deblurs} and \red{super-resolves} the depth map by using the second-order TGV model.} \\
\cline{2-4}
~&FCN-PDN~\cite{riegler2016deep} & BMVC-2016 & \thead[l]{Models the TV regularizer in a non-local neighborhood and learns the weighting factor via a network.}\\
\cline{2-4}
~ & NBTV~\cite{wang2016super} & TIP-2016& \thead[l]{Integrates the bilateral filter~\cite{BF} into the TV to preserve the sharpness and consistency of edges.} \\
\hline
\multirow{19}*{Garph Laplacian}& JGIE~\cite{8491336} & TIP-2019& \thead[l]{Proposes a novel GDSR method by combining two graph domain regularizations: the internal\\ smoothness prior and the external gradient consistency constrain.} \\
\cline{2-4}
~& JARMT~\cite{8474366}& TIP-2019&\thead[l]{Proposes a depth map restoration method which includes a local and a non-local manifold models to\\ fully exploit the local smoothness prior and the non-local self-similarity structures. } \\
\cline{2-4}
~& ADFTG~\cite{8598824}& TCSVT-2020 & \thead[l]{Proposes an adaptive data fidelity term to optimally generate each depth pixel and an unified graph\\-based regularization term to preserve the piecewise smooth and sharpen the edges.} \\
\cline{2-4}
~ &GNNLG~\cite{yan2020depth} & TOMM-2020 & \thead[l]{Proposes a model which combines the nuclear norm and group-based graph prior to exploit the low\\-rank and non-local similarity properties of the depth map.} \\
\cline{2-4}
~&DSSR~\cite{9511489} & TCSVT-2021 & \thead[l]{Proposes a dual normal-depth regularizer to exploit the geometric relationship between the depth and\\ its corresponding normal and a reweighted graph Laplacian regularizer to obtain sharp edges.} \\
\cline{2-4}
~& GraphSR~\cite{GraphSR} & CVPR-2022 &\thead[l]{Proposes to learn the affinity graph and then embeds the learned graph to a differentiable optimisation\\ layer to regularize the upsampling process.} \\ 
\hline
\multirow{20}*{Others} & LR~\cite{Lu} & CVPR-2014 & \thead[l]{Proposes a low-rank matrix completion-based depth map enhancement method which is robust to the\\ noise and weak correlation between the depth map and its corresponding color image..}\\
\cline{2-4}
~ & JLNLR~\cite{7576720} & TMM-2017 & \thead[l]{Combines both local and non-local regularization for depth map recovery. The local regularization \\ contains a weighted TV and a dual AR model, and the non-local regularization is a low-rank constrain.}\\
\cline{2-4}
~ & RGDMR~\cite{7574299} & TIP-2017 & \thead[l]{Proposes a robust penalty function as smoothness term to handle the issue of structure inconsistency.}\\
\cline{2-4}
~ & SPIDM~\cite{ye2020sparsity} & PR-2020 & \thead[l]{Proposes a image decomposition model for depth map recovery, where the depth map is decomposed\\ into a local smooth surface and a piecewise constant component. The former is modeled by a least\\ square polynomial approximation, while the latter is modeled by a sparsity-promoting prior.} \\
\cline{2-4}
~& DGDIE~\cite{gu2019learned} & TPAMI-2019 & \thead[l]{Proposes a weighted analysis representation model for depth map reconstruction. The optimization\\ process of this model is unfolded as a series of stage-wise operations which can be obatined by a\\ task-driven training strategy.} \\
\cline{2-4}
~& GSF~\cite{liu2021generalized}& TPAMI-2021& \thead[l]{A generalized framework which combines the truncated Huber penalty function for image filtering. \\By varying the parameters of this function, the framework can obtain various smoothing natures.} \\
% \hline
% \multirow{6}*{Hybrid Model}  & TSDR~\cite{jiang2018depth} & TIP-2018 & \thead[l]{A method which includes both transform  and spatial domain\\ regularizers. The transform domain regularizer is a patch-based\\ AR model to characterize intra-pacth correlations and the spatial\\ domain regularizer is a multi-directional TV to capture the \\geometrical structures.} \\
% \cline{2-4}
% ~ & SRAM~\cite{wang2019multi} & TMM-2019 &  \thead[l]{A method which combines multi-direction sparse representation\\  and AR  to represent depth edges in both patch and pixel level. }\\
% ~& DGDIE~\cite{gu2019learned} & TPAMI-2019 &  \\
\bottomrule
\end{tabular}
\end{center}
\vspace{-0.2in}
\end{table}

\subsection{Auto-regressive Model}
The center theme of Auto-regressive (AR) Model is that a signal can be represented as a linear combination of itself with carefully modified AR coefficients. Let $\bm{y}\in \mathbb{R}^N$ be the input signal, the output of AR model at position $p$ is given by:
$
    \bm{y}^*_p = \sum_{q\in \mathcal{N}_p}\alpha_{p, q}\bm{y}_q,
    %\label{eq_ar_1}
$
where $\mathcal{N}_p$ means the neighborhood pixel of the pixel $p$, and $\alpha$ are the AR coefficients. \citet{6827958} find that the depth signals can be well represented by the AR model if we can adjust the coefficients based on the unique characteristics of the depth signals. To this end, the authors propose an adaptive AR model for GDSR which adjusts the coefficients by considering both the local correlation in the LR depth map and the non-local similarity in the HR guidance image. This model is given by:
\begin{equation}\small
    \argmin_{\bm{y}^*} \sum_{p \in \mathcal{O}} (\bm{y}^*_p - \hat{\bm{y}}_p)^2 + \sum_p \left(\bm{y}^*_p - \frac{1}{C_p}\sum_{q \in \mathcal{N}_p}\exp{\left(-\frac{(\bm{y}^*_{p} - \bm{y}^*_q)^2}{2\sigma_1^2}\right)}\exp{\left(-\frac{\sum_{i\in \mathcal{C}}\left\| \mathbf{B}_p \circ\left(\mathcal{P}_p^i - \mathcal{P}_q^i\right) \right\|_2^2}{2\times 3 \times \sigma_2^2}\right)} \bm{y}^*_q\right),
\end{equation}
where $\mathcal{O}$ denotes the set of pixels of the coarsely interpolated depth map $\hat{\bm{y}}$, and $\mathcal{N}_p$ means the neighborhood pixel of the pixel $p$; $\sigma_1$ and $\sigma_2$ are constant values to control the decay rate of the exponential function; $\mathcal{P}_p^i$ represents the operator used to extract a patch centered at $p$ in the color channel $i$; $\circ$ is pixel-wise multiplication; $\mathbf{B}$ denotes the bilateral kernel function that is used to capture shape information of local image structures. \citet{jiang2018depth} find that the pixel-based AR model tends to produce undesired artifacts when there is a large difference between the center pixel and its neighbors. To remedy this limitation, the authors propose a patch-based AR model where the relationship is modeled by non-local similar patches. 
% The pixel-based AR model in Eq.~\ref{eq_ar_1} changes to the patch-based AR model if we set the $\mathcal{N}_p$ as the set of similar patches of patch $\bm{y}^*_p$. They further propose to represent the patch in the transform domain and enforce the patch coefficient to be sparse, thus the influence of retrieved similar patches that have different details can be mitigated. 
\citet{wang2019multi} propose a compound method by combining dictionary learning and the AR model. The dictionary learning is used to represent the inter-patch (patch-level) correlation of depth edges, while the AR model is used to describe the intra-patch (pixel-level) correlation of depth edges. \citet{wang2016super} extend the AR model by selecting the largest variance channel as input, as they find that the average operator adopted in the original AR model tends to impair structural awareness.

\subsection{Total Variation}
Total variation (TV)~\cite{rudin1992nonlinear} or ROF model is a classical regularization functional proposed by Rudin et al. in 1992, and has been widely used in the area of image restoration. The TV~\cite{rudin1992nonlinear} distinguishes the edge and the noise using the image gradients, and diffusion is performed only along the edge direction; thus the edge information can be well preserved. Unfortunately, the TV can only approximate the piecewise constant function and tends to generate the “staircasing artifacts” for the smooth region. To solve this problem, \citet{bredies2010total} propose the Total Generalized Variation (TGV), a generalized mathematical model of TV that can approximate the piecewise multinomial function of any order. Motivated by this, several methods have been proposed that incorporate TV and its variants as regularizers and achieved promising performance for the GDSR.

\citet{ferstl2013image} propose to use a second-order total generalized variation (TGV) weighted by an anisotropic diffusion tensor as the regularization function to upsample the depth map, which can avoid surface flattening. The energy function is formulated as follows:
\begin{equation}
    \min_{\bm{y}^*, \bm{z}} \|\bm{y}^* - \hat{\bm{y}}\|^2_2 + \alpha_1 \bm{T} \lvert \nabla \bm{y}^* + \bm{z} \rvert +\alpha_0 \lvert \nabla \bm{z}\rvert,
\end{equation}
where the last two terms of this equation denote the second-order TGV regularizer; $T$ is the anisotropic diffusion calculated from the HR guidance image.
%to weight the depth gradient and guide the direction of the gradient during the optimization process. 
% The tensor is given by:
% \begin{equation}
%     \bm{T} = \exp{(-\beta \lvert \nabla \bm{g} \rvert^\gamma)\bm{n}\bm{n}^\mathrm{T}} + \bm{n}^{\perp} \bm{n}^{\perp \mathrm{T}},
% \end{equation}
% where $\bm{n}$ means the normalized gradient of the guidance image $\bm{n}=\frac{\nabla g}{\lvert \nabla g \rvert}$; $\bm{n}^{\perp}$ means the normal vector of the gradient; $\beta$ and $\gamma$ are two scalars that are used to rectify the magnitude and sharpness of the tensor, respectively. To facilitate the quantitative comparison of GDSR algorithms, the authors also propose a real ToF dataste which is equipped with the high-resolution ground truth depth maps. 
\citet{riegler2016atgv} propose to incorporate the second-order TGV model into a deep convolution network, which combines the merits of deep learning and variational methods. In this model, all the optimization procedures of the variational model are unrolled so that the whole model can be trained end-to-end. \citet{riegler2016deep} propose a non-local TV (NLTV) regularization with a robust norm for GDSR. The objective function of NLTV is formulated as follows:
\begin{equation}
    E = \sum_p \sum_{q\in \mathcal{N}_p}  \exp{\left(-\frac{\|p-q\|_2}{\sigma_d} - \frac{\|\hat{\bm{g}}_p -\hat{\bm{g}}_q\|_2}{\sigma_v}\right)} \lvert \bm{y}^*_p - \bm{y}^*_q\rvert_{\epsilon},
\end{equation}
where $\lvert \cdot \rvert_\epsilon$ denotes the Huber norm\cite{huber1973robust} which can not only smooth the depth reconstruction, but also preserve sharp boundaries. $\epsilon$ is the parameter for the Huber norm; $\hat{\bm{g}}$ is the guidance image learned by a neural network. 
%\citet{xiao2015defocus} propose to apply the second-order TGV to the latent amplitude and the depth map to simultaneously defocus deblurring and super-resolution. 
\citet{wang2016super} propose the normalized bilateral TV (NBTV) which extends the TV by involving the bilateral filter~\cite{BF} and the normalized sparse measure. Compared to the TV, the NBTV~\cite{wang2016super} can produce results with sharper and more precise edges. The edges of the depth map are typically oriented at arbitrary directions. Nevertheless, most of the existing TV-based models only deal with the horizontal and vertical directions, which may oversmooth the fine details. \citet{jiang2018depth} propose a weighted multi-direction TV (WMTV) that represents depth edges with spatially varying orientations. The objective function of WMTV is as follows:
\begin{equation}
    E=\sum_i\|\bm{\mathcal{W}}_i \nabla_{\theta_i} \bm{y}^* \|_1,
\end{equation}
where $\nabla_{\theta_i}$ means the gradient along the direction $\theta_i$ and $\bm{\mathcal{W}}_i$ denotes a weighting matrix that is used to assign different values to different directions for each pixel. 

\subsection{Graph Laplacian}
One of the most significant image priors for depth map restoration is that depth maps contain smooth areas separated by sharp boundaries, which is also known as piecewise smooth (PWS). Recent studies have shown the powerful capability of the graph Laplacian to deal with the PWS signal. Inspired by this, several works have been proposed to use it to exploit the smoothness of the depth map. In the following, we first present the general formulation of the graph Laplacian and then introduce some representative methods that employ the graph Laplacian as the regularizer.

A depth map can be modeled as a signal on a weighted undirected graph $\mathcal{G}=(\mathcal{V}, \mathcal{E}, \bm{W})$, where $\mathcal{V}$ denotes the set of vertices (or nodes); $\mathcal{E}$ means the set of edges, and each edge connects a pair of vertices in $\mathcal{V}$;  $\bm{W}$ denotes the weighted adjacent matrix that is used to encode the similarity between two connected vertices. Note that we only consider $\mathcal{G}$ as an un-directed graph with non-negative edge weights, so we have $\bm{W}_{ij}=\bm{W}_{ji}$ and $\bm{W}_{ij}>0$. The graph Laplacian matrix $\bm{L} = \bm{D} - \bm{W}$,
where $\bm{D}_{ii}=\sum_j \bm{W}_{ij}$ means the diagonal degree matrix. Given the graph Laplacian matrix $\bm{L}$, the smoothness of a signal $\bm{y}$ on the graph can be formulated as a quadratic form of $\bm{L}$:
\begin{equation}
    \bm{y^\top\bm{L}\bm{y}} = \frac{1}{2}\sum_{i,j} \bm{W}_{ij} (\bm{y}_i - \bm{y}_j)^2,
    \label{gh_1}
\end{equation}
by minimizing Eq.~\ref{gh_1}, the signal $x$ is promoted to be smooth with respect to the graph $\mathcal{G}$.

Generally, there are two main graph-based strategies used in the literature for the inverse problem: graph-based regularization and graph-based transform. For the graph-based regularization, the signal is generally assumed to be smooth with respect to the graph $\mathcal{G}$, and the image restoration task can be solved by the following equation:
\begin{equation}
    \bm{y}^* = \argmin_{\bm{y}^*} \frac{1}{2} \|\bm{y} - \bm{H}\bm{y}^* \|_2^2 + \gamma \bm{y^\top\bm{L}\bm{y}^*},
    \label{eq_gh_2}
\end{equation}
the regularization term is small if each pair of nodes $(i, j)$ which are connected by an edge has similar values or the edge weight $\bm{W}_{ij}$ is small. 
% The closed-form solution of Eq.~\ref{eq_gh_2} is given by:
% \begin{equation}
%     \bm{y}^* = \left(\bm{H^\top\bm{H} + \gamma \bm{L}}\right)^{-1}\bm{H}^\top \bm{y}.
% \end{equation}
Since the Laplacian matrix $\bm{L}$ a real, symmetric, positive semi-definite matrix, we can decompose it into a set of orthogonal eigenvectors, denoted by $\{\bm{u}_l\}_{l=1,\cdots, n}$, with real non-negative eigenvalues $0=\eta_1 \leq \eta_2 \leq \cdots \leq \eta_n$:
\begin{equation}
    \bm{L}=\bm{U}\Lambda\bm{U}^\top,
\end{equation}
where $\bm{U}$ is the eigen-matrix with $\bm{u}_i$'s as columns and $\Lambda$ is the diagonal matrix with $\eta_i$'s on its diagonal. The graph-based transform (GFT) can be formulated as: $\alpha = \bm{U}^\top \bm{y}$.
% \begin{equation}
%     \alpha = \bm{U}^\top \bm{y}.
% \end{equation}

The graph-based regularization can achieve better performance for the image super-resolution task, while the graph-based transform works well for the image denoising task~\cite{8598824}. To this end, \citet{8598824} propose a unified regular function named transferred graph regularization for image restoration tasks, which combines the advantages of the above two graph-based strategies. 
% Moreover, they also propose a new data fidelity term based on a mixture probability maximization model where the mixture coefficients are obtained by the adaptive AR model proposed in \cite{6827958}. 
\citet{8491336} propose a novel framework for GDSR that contains an internal smooth prior and an external gradient consistency constraint. For the internal smooth prior, the authors propose a graph Laplacian regularizer where the weight matrix is tailored to fully utilize information between the depth map and its corresponding color image. The external gradient consistency constraint is proposed to address the problem of structure discrepancy, and this constraint is motivated by the observation that the depth map gradient is small apart from the edge-separating region. Hence, the depth map gradient can be obtained by discarding the finite parts of the guidance image gradient:
\begin{equation}
    \nabla_\mathcal{G} \bm{y} = \mathcal{T}_{\tau}(\nabla_\mathcal{G} \mathbf{g}),
\end{equation}
where $\mathcal{T}_{\tau}$ denotes the hard thresholding operator with parameter $\tau$; $\nabla_\mathcal{G} \bm{y}$ and $\nabla_\mathcal{G} \mathbf{g}$ mean the gradients of the depth map and the guidance image, and the gradient operator is defined as follows:
\begin{equation}
   \nabla_\mathcal{G} \bm{y}_i=\sum_{j, j \sim i} \bm{W}(i, j)(y(i)-y(j)),
\end{equation}
where $\bm{W}_{ij}$ means the edge weight between nodes $i$ and $j$; $j \sim i$ denotes that node $j$ is connected to node $i$ in graph $\mathcal{G}$. The final object function of this work is given by:
\begin{equation}
    \argmin_{\bm{y}^*} \|\bm{y}-\bm{Hy}^* \|_2^2 + \alpha \bm{y}^{*\top} \hat{\bm{L}}\bm{y}^* + \beta \|\nabla_\mathcal{G} \bm{y}^* - q\|_1,
    \label{eq_l_1}
\end{equation}
where $\hat{\bm{L}} = \bm{LD}^{-1}\bm{L}$; $q$ is a constant value which means the threshold of the guidance image gradient. The second term in Eq.~\ref{eq_l_1} is the internal smoothness prior, while the third term in Eq.~\ref{eq_l_1} means the external consistency constraint. \citet{8474366} propose a depth map restoration method which includes local and non-local manifold models. The local manifold is used to maintain the smoothness of the local manifold structure, and the non-local manifold model is designed to make full use of the self-similarity structures to build highly data-adaptively orthogonal basis. 
% In addition, the authors propose a 3D thresholding operator to make the decomposition of the signal on the %manifold spectral bases sparse; thus, only the low graph frequencies of the depth map can be preserved. 
Instead of constructing the graph with the pixels, \citet{yan2020depth} propose to construct the graph with a group of similar patches, as the group-based image restoration model can efficiently gather the similarity of different patches. 
%The similar patches are obtained by the block matching algorithm. 
%They further propose a nuclear norm to exploit the low-rank property of the group. 
\citet{9511489} find that the edge weight %($d=\lvert\bm{y}^*_i -\bm{y}^*_j \rvert$) 
distribution of an area with sharp edges is a bimodal distribution. Inspired by this, they propose a reweighted graph Laplacian regularizer to preserve sharp edges and promote the bimodal distribution of edge weights. 
% The new Laplacian regularization is given by:
% \begin{equation}
%     \bm{y}^{*\top} \bm{Lx} = \frac{1}{2}\sum_i \sum_j (\bm{y}^*_i-\bm{y}^*_j)^2 \bm{C} \exp{\left(-(\bm{y}^*_i-\bm{y}^*_j)^2 / (2\times \sigma_d^2) \right)},
% \end{equation}
% where $\bm{C}=\exp{\left(-\|\sum_{c\in{R,G,B}}\bm{I}_c(i) - \bm{I}_c(j)\| \right /3\times \sigma^2)}$ means the proposed weighting term and $\bm{I}$ is the guidance image. Furthermore, they propose a dual normal-depth regularization to make full use of the geometric features contained in the normal map. 
To fully exploit the contextual information of the guide image and ensure strict fidelity of the super-resolved depth map to the low-resolution one, \citet{GraphSR} propose to learn the graph regularization. Specifically, they first use a deep neural network to learn the affinity graph; then, the learned affinity graph is sent to a differentiable optimization layer to regularize the upsampling process.

\subsection{Others}
In this subsection, we introduce some methods that employ optimization algorithms to address the ill-posed GDSR problem but do not belong to any of the above categories.

In the literature, the GDSR methods are mainly based on two assumptions: 1) the guidance image should be of high quality, 2) the depth map and its corresponding guidance image should be highly correlated. However, these two assumptions may be invalidated in some cases. For example, the guidance image is usually noisy when the depth camera is used in low light conditions. The noisy image is prone to mislead the depth restoration algorithm and pollute the depth edges. To overcome these problems, \citet{Lu} propose a low-rank matrix completion-based method that takes into account the noisy guidance image. Their method is based on the observation that similar RGB-D patches lie in a very
low-dimensional subspace. By assembling similar patches into a matrix and restricting low-rank constraint, the latent structure structure in the RGB-D patches can be essentially captured. 
%Moreover, the authors also propose a learning method for automatic rank estimation.
\citet{7576720} propose to combine both local and non-local regularization strategy for the depth map reconstruction. The local regularization contains two complementary local constraints: the gradient domain weighted TV and the spatial domain dual AR model. The non-local regularization is a low-rank constraint which aims to exploit the global characterization of color-depth dependency. Instead of designing an effective guidance weight, \citet{7574299} propose a robust penalty function as a smoothness term to alleviate texture-copying artifacts.
% \begin{equation}
%     \argmin_{\bm{y}^*} \|\hat{\bm{y}}-\bm{y}^*\|_2^2 + \alpha \sum_i \sum_{j\in \mathcal{N}_i} \exp{\left(-\frac{\lvert i-j\rvert^2}{2\sigma_s^2}\right)} \exp{\left(-\frac{\lvert\bm{g}_i -\bm{g}_j \rvert^2}{3\times 2\sigma_c^2}\right)} \psi_\mu(\lvert \bm{y}^*_i -\bm{y}^*_j \rvert^2), 
% \end{equation}
% where $\psi_\mu(x^2)=2\mu^2(1-\exp{(-x^2/2\mu^2)})$ denotes the proposed robust penalty function; $\mu$ is a parameter, and the authors introduce a data-driven method to estimate this parameter.
\citet{ye2020sparsity} find that the depth map is mainly consists of smooth regions separated by sharp edges, and propose a image decomposition model for depth map recovery. Concretely, the authors first decompose the depth map into a local smooth surface and an approximately piecewise constant component. Then, they propose using a polynomial smoothing filter to approximate the local smooth surface and a sparsity-promoting prior to model the piecewise constant component. 
% Moreover, they propose an iterative reweighted strategy to address the structural inconsistency problem. 
\citet{gu2017learning, gu2019learned} propose a weighted analysis representation model (WASR) for guided depth map reconstruction, in which they combine the analysis representation regularization term and the guidance weighting term to model the relationship of the RGB-D pairs. The optimization of WASR is unfolded; thus, the proposed model can make full use of previous expertise as well as training data. 
% In addition, they propose to smooth depth maps by layer decomposition and adopt different layers for different sparse representations. 

% \citet{Liu_Zhang_Lei_Huang_Yang_Reid_2020, liu2021generalized} propose a general image filtering framework, which combine the truncated Huber penalty function. By varying the parameters of this function, the framework can obtain various smoothing natures. Moreover, the authors propose an efficient numerical solution to solve this nonconvex and nonsmooth optimization framework.

\section{Learning-based Methods}\label{learn}
\subsection{ Dictionary Learning Methods}
In this section, we introduce the learning-based methods, which include the dictionary learning methods and the deep learning methods. The dictionary learning methods aim at finding a set of basis atoms (dictionary) such that the input signal can be represented as a sparse combination of these atoms. Contrary to dictionary learning methods, the deep learning methods directly learn the non-linear mapping from the two inputs (i.e., the guidance image and the low-resolution depth map) to the ground truth depth map by using the deep neural network. A brief summary of the learning-based methods can be found in Tab.~\ref{tab_learning}.
\begin{table}[!t]\setlength{\tabcolsep}{2.5pt}\renewcommand{\arraystretch}{0.8}\renewcommand\theadfont{\tiny}
    \tiny 
    \begin{center}
    \vspace{-0.14in}
    \caption{\footnotesize A brief summary of the learning-based methods.}
    \vspace{-0.09in}
    \label{tab_learning}
    \begin{tabular}{lccc}
\toprule
Category & Methods & Published &  Basic Idea \\
\midrule
\multirow{28}*{Dictionary Learning} & JIDSR~\cite{6775324}  &  TIP-2014 & \thead[l]{Proposes a model to
learn joint sparse representations of the depth and the guidance images. In the model,\\ the dictionaries and the coefficients for each image are different, but share a common sparse support. \\Moreover, it proposes a Joint Basis Pursuit algorithm to infer the sparse coefficients.} \\
\cline{2-4}
~& CDLSR~\cite{song2019multimodal} & TCI-2019 & \thead[l]{Proposes a coupled dictionary learning method to explicitly capture the multimodal dependency in the sparse\\ feature domain. To better characterize the similarities and disparities between different modalities, the\\ authors propose to use common and unique sparse representations to represent the multimodal images.}\\
\cline{2-4}
~& RADAR~\cite{8741062} & TCSVT-2019 & \thead[l]{Proposes a method which combines the strengths finite rate of innovation (FRI) theory with multimodal\\ dictionary learning for robust depth map super-resolution.The FRI theory is used to do initial upsampling,\\and the multimodal dictionary learning is used to capture meaningful information from the guidance image.}\\
\cline{2-4}
~& JMDL~\cite{8858035}& TIP-2019& \thead[l]{Proposes a joint multimodal dictionary learning algorithm that simultaneously learn three dictionaries and\\ two transform matrices. The learned dictionaries are used to represent the LR depth map, the guidance\\ image, and the HR depth map, while the learned transform matrices are employed to model the relationship\\ among the spare representations of each modality.} \\
\cline{2-4}
~& LMCSC~\cite{marivani2020multimodal} & TIP-2020 & \thead[l]{Proposes a multimodal sparse coding model where different modalities are described by different\\ dictionaries, and the same sparse representations for each modalities are assumed to be similar.} \\
\cline{2-4}
~&CUNet~\cite{CUNet} & TPAMI-2021 & \thead[l]{Proposes to represent each modality with two convolutional dictionaries, where one dictionary is utilized to\\ represent the common features and the other is utilized to represent unique features.}\\
\cline{2-4}
~&MCDL~\cite{9681224} & TIP-2022 & \thead[l]{Proposes to solve the multimodal convolution dictionary learning problem by traditional optimization, which\\ is demonstrate to be less sensitive than the deep unfolding strategy with limited training data.} \\
\hline
\multirow{60}*{Deep Learning} & DMSG~\cite{DMSG} & ECCV-2016 & \thead[l]{Proposes a multi-scale guided convolution network with multi-scale fusion strategy and high-frequency\\ domain
training method to boost the performance of depth map reconstruction.}\\
\cline{2-4}
~& MGD~\cite{zuo2019depth} & TCSVT-2019 &\thead[l]{Proposes a method to improve the effectiveness of the guidance features. The method extracts the multi-scale\\ guidance features in the high-resolution domain and revisits the extracted features via dense connection.}\\
\cline{2-4}
~ & DJFR~\cite{DJFR} & TPAMI-2019 & \thead[l]{Proposes a two-branch network with skip connection for joint image filtering.}\\
\cline{2-4}
~& CCFN~\cite{CCFN} & TIP-2019& \thead[l]{Proposes a coarse-to-fine convolution neural network for depth map super-resolution. In the coarse stage,\\ the authors use large filter kernels to obtain crude high-resolution depth, and in the fine stage, the authors\\ use small filter kernels to produce more accurate results.}\\ 
\cline{2-4}
~& DepthSR~\cite{DepthSR} & TIP-2019& \thead[l]{Proposes a residual U-Net architecture with hierarchical guidance branch and input pyramid branch.}\\ 
\cline{2-4}
~& DSRN~\cite{DSRN} & PR-2020 &\thead[l]{Proposes a novel framework which is based on deep edge-aware learning. This framework firstly learns\\ the edge information, and then recovers the HR depth map by using the proposed fast depth filling\\ module and cascaded network.} \\
\cline{2-4}
~& CCMSN~\cite{li2020depth} & PR-2020 &\thead[l]{Proposes a multi-scale symmetric network for depth map super-resolution, and it's core is a correlation\\-controlled color guidance block which aims to improve the color guidance accuracy.} \\
% ~& MFR~\cite{8598786} & TCSVT-2020& \thead[l]{A multi-scale framework which contains local and global residual learning and batch normalization. The\\ residual learning is used to ease the network training, while the batch normalization is used to improve\\ the performance of depth map denoising.} \\
\cline{2-4}
~&CGN~\cite{CGN} & TMM-2020& \thead[l]{Proposes a coarse to fine framework which contains a depth-guided intensity features filtering module and\\ a intensity-guided depth features refinement module to mitigate the artifacts caused by edge misalignment.}\\
\cline{2-4}
~& PMBAN~\cite{PMBAN}& TIP-2020 & \thead[l]{Proposes a progressive multi-branch aggregation network (PMBAN), which contains stacked MBA blocks\\ to progressively reconstruct the depth map. The MBA block has multiple branches to extract multimodal\\ features, and the extracted feature are fused by the proposed fusion block.} \\
\cline{2-4}
~& MIG~\cite{MIG} & TMM-2021 & \thead[l]{Proposes a multi-stage network to enhance the LR depth map. In each stage, the depth features in the image\\ and the gradient domains are iteratively refined by the guidance features through two parallel streams.}\\
\cline{2-4}
~& AHMF~\cite{AHMF}& TIP-2021 &\thead[l]{Proposes a novel model which contains a multimodal attention based fusion module and a bidirectional\\ hierarchical feature collaboration module to fuse multimodal and multilevel complementary features.} \\
\cline{2-4}
~& JIIF~\cite{JIIF} & MM-2021 & \thead[l]{Models the GDSR as a neural implicit interpolation process, and the interpolation weights and values are\\ learned by
a proposed novel joint implicit image function.}\\
\cline{2-4}
~ &Bridge~\cite{tang2021bridgenet}&MM-2021& \thead[l]{Proposes a multi-task learning framework that combines the depth map super-resolution and monocular\\ depth estimation to improve the quality of reconstructed depth maps. The two tasks are interacted with\\ a high-frequency bridge and a context guidance bridge.}\\
\cline{2-4}
~& CMSR~\cite{shacht2021single}& CVPR-2021& \thead[l]{Proposes a training strategy for GDSR, which only need one weakly aligned RGB-D pair as the training data.} \\
\cline{2-4}
~& FDSR~\cite{FDSR} & CVPR-2021 & \thead[l]{Proposes a high-frequency guidance network which employs the octave convolution~\cite{chen2019drop} to capture the high-\\frequency components of the RGB image to guide the depth map super-resolution process.}\\
\cline{2-4}
~& CTCK~\cite{sun2021learning}& CVPR-2021 &\thead[l]{Proposes to utilize the guidance information by knowledge distillation. In this method, the guidance image\\ is only required at the training stage.} \\
\cline{2-4}
~ & DCTNet~\cite{DCTNet} & CVPR-2022 &\thead[l]{Proposes a method that extends the discrete cosine transform to a deep learning module to make the network\\ explainable, and combines this module with a shared/private feature extraction and guided edge attention\\ mechanism to address issues of cross-modal feature extraction difﬁculty and RGB texture over-transferred.} \\
\bottomrule
\end{tabular}
\end{center}
\end{table}

In this subsection, we introduce the  dictionary learning methods. We first present the general formulation of the dictionary learning, and then introduce some representative methods that employ the dictionary learning to solve the ill-posed guided depth map super-resolution (GDSR) problem. Note that the convolution dictionary learning methods are also included in this subsection.

Generally, the sparse dictionary learning consists of two steps: sparse coding and dictionary learning. Let $\bm{y} \in \mathbb{R}^N$ be an input signal and $\bm{D} \in \mathbb{R}^{N\times M} (N \ll M)$ be an over-complete dictionary; the sparse coding process aims to find a representation of $\bm{y}^*$ in the form of a linear combination of a small number of atoms in $\bm{D}$, i.e., $\bm{y}^*\simeq \bm{D\alpha}$ where $\bm{\alpha} \in \mathbb{R}^M$ should be sparse. The process of finding the sparse representation $\bm{\alpha}$ can be formulated as the following optimization problem:
\begin{equation}
    \argmin_{\alpha} \|\bm{y}^* - \bm{D\alpha}\|_2 + \|\bm{\alpha}\|_0,
    \label{eq_dic_1}
\end{equation}
where $\|\cdot\|_0$ means the $l_0$ pseudo-norm counting the number of nonzero elements in the vector $\bm{\alpha}$. The above minimization problem is not convex due to the $l_0$ norm, and solving this problem is NP-hard. There are two typically ways to solve the above problem, one is to employ greedy methods such as orthogonal matching pursuit (OMP)~\cite{tropp2007signal}, the other is to relax the $l_0$ norm to the $l_1$ norm and solve it by using basis pursuit~\cite{chen2001atomic} or LASSO~\cite{santosa1986linear}. The dictionary learning step aims to find an over-complete dictionary such that the fidelity term in Eq.~\ref{eq_dic_1} can be minimized when the $\bm{\alpha}$ is fixed. In the literature, many approaches have been proposed to learn the dictionary, such as MOD~\cite{760624}, K-SVD~\cite{aharon2006k} and Lagrange dual method~\cite{lee2006efficient}.

Sparse dictionary learning has long been employed to represent images and signals. To represent multi-modal images, \citet{6775324} propose a sparse generative model to learn the joint representation of the depth map $\bm{y}^*$ and the guidance image $\bm{g}$, where they represent different modalities with different dictionaries and different coefficients:
\begin{equation}
    \bm{g} =\bm{D}^g \bm{a} + \bm{\eta}^g, \quad \bm{y}^* = \bm{D}^y \bm{b}+ \bm{\eta}^y,
\end{equation}
where $\bm{\eta}$ is the noise vector; $\bm{a}=[a_1, \cdots, a_K]$ and $\bm{b}=[b_1, \cdots, b_K]$ are the sparse coefficients for the guidance and depth images, respectively. As the depth and guidance images are captured from the same scenes, the authors couple these two sets of sparse coefficients via latent variables $\phi_i$, which can be defined as follows:
\begin{equation}
    a_i = m^g_i \phi_i,\quad b_i = m^x_i \phi_i,\end{equation}
where $m_i^g$ and $m_i^x$ denote the magnitudes of the sparse coefficients, and $\phi_i$ represents the activity of the $i-$th sparse coefficient. By restricting the vector $\bm{\phi}=[\phi_1, \cdots, \phi_K]$ to be sparse, we can obtain the sparse representation of both the depth and the guidance images. 
% In addition, the authors propose a second-order cone algorithm named Joint Basis Pursuit as a replacement for the traditional Group Lasso algorithm to simultaneously estimate the latent variables $\bm{\phi}, \bm{m}^g$ and $\bm{m}^x$.
To better reconstruct the depth edges, \citet{wang2019multi} propose a multi-direction dictionary learning algorithm, where they first classify the patches according to the geometrical directions, and then train the dictionaries for each class, respectively. \citet{song2019multimodal} propose a coupled dictionary learning based multi-modal image super-resolution (CDLSR) method to exploit complex dependencies between different modalities in a learned sparse feature domain. The CDLSR~\cite{song2019multimodal} is made up of a coupled dictionary learning phase and a coupled super-resolution phase. The learning phase aims at learning a set of dictionaries to couple the multi-modal images together in the sparse feature domain, and can be formulated as follows:
\begin{equation}
{\footnotesize \argmin_{\begin{array}{c}
\{\boldsymbol{\Psi}_{c}^{l}, \boldsymbol{\Psi}^{l}, \boldsymbol{\Psi}_{c}^{h},
\boldsymbol{\Psi}^{h}, \boldsymbol{\Phi}_{c}, \boldsymbol{\Phi}\} \\
\{\bm{Z}, \bm{U}, \bm{V}\}
\end{array}}}
    \left\|\left[
    \begin{array}{c}
        \bm{Y}^{l} \\
        \bm{Y}^{*} \\
        \bm{G}
    \end{array}
    \right]-\left[
    \begin{array}{ccc}
        \bm{\Psi}_{c}^{l} & \bm{\Psi}^{l} & \bm{0} \\
        \bm{\Psi}_{c}^{h} & \bm{\Psi}^{h} & \bm{0} \\
        \bm{\Phi}_{c} & \bm{0} & \bm{\Phi}
    \end{array}\right]\left[
    \begin{array}{c}
        \bm{Z} \\
        \bm{U} \\
        \bm{V}
    \end{array}\right]\right\|_{F}^{2} \text{s.t.} \left\|\bm{z}_{i}\right\|_{0}+\left\|\bm{u}_{i}\right\|_{0}+\left\|\bm{v}_{i}\right\|_{0} \leq s, \forall i,
\end{equation}
where $\bm{Y}^l, \bm{Y}^*$ are the low-resolution (LR) and the high-resolution (HR) depth maps, respectively; $\bm{G}$ is the guidance image; $\bm{\Psi}$ and $\bm{\Phi}$ represent the learned dictionaries and $\bm{Z, U}$ and $\bm{V}$ represent the sparse representations; $\|\cdot\|_F$ is the Frobenius norm. In order to better characterize the similarities and disparities between different modalities, the authors make the sparse representations $\bm{Z}$ and $\bm{U}$ shared for the same modality, and the sparse representations $\bm{U}$ and $\bm{V}$ different for the different modalities. To further improve the performance of this method, the authors propose to incorporate the multistage strategy and the neighborhood regression technique into CDLSR~\cite{song2019multimodal}. \citet{8741062} propose a robust algorithm for guided depth map super-resolution, named RADAR, which combines the strengths of finite rate innovation (FRI) theory and multi-modal dictionary learning. 
% Given the LR depth map as input, the authors first model its rows and columns as piecewise signals and propose a FRI-based algorithm to upsample it to a moderate quality (MQ) depth map. Then, they send the MQ depth to the proposed projection-based rapid upscaling algorithm which is based on a multi-modal dictionary learning model~\cite{song2019multimodal} to further improve the reconstruction performance. 
\citet{8858035} propose a joint multi-modal dictionary learning (JMDL) algorithm for GDSR. In JMDL the authors use three dictionaries to represent the LR depth map, the guidance image, and the HR depth map, respectively. Considering that the three images are captured from the same scene, the authors assume that the sparse representations of them are related. 
% Let $\bm{a}, \bm{b}$ and $\bm{c}$ be the representation for the LR depth map, the guidance image, and the HR depth map, respectively, and the relationship among them is given by the following equation:
% \begin{equation}
%     \bm{c}=S_\lambda (\bm{W}_a \bm{a} + \bm{W}_b \bm{b}),
% \end{equation}
% where $S_\lambda (\cdot)$ denotes the soft-thresholding operator and $\bm{W}$ is the transform matrix.

To guarantee the sparse representation, the dictionary $\bm{D}$ should be sufficiently over-complete. This makes the sparse coding model only represent the low-dimensions signals and may suffer from curse of dimensionality for the high-dimensions signals, such as image data. To address this problem, the researchers propose to split the image into overlapping patches and perform sparse coding on each patch. However, this mechanism neglects the dependencies between patches, which tends to produce unsatisfactory results. To overcome these shortcomings, the convolutional sparse coding model~\cite{bristow2013fast, wohlberg2015efficient} has been proposed, which employs convolution filters to replace traditional dictionary atoms. The convolutional sparse coding model is spatially invariant and avoids splitting the image into patches; thus, the consistency prior can be naturally exploited. Given an input image $\bm{y}$ and dictionary filters $\bm{D}=\{\bm{d}_i | i=1, \cdots, N\}$, the goal of the convolution sparse coding model is to calculate a set of convolution sparse maps $\bm{R}=\{\bm{\alpha}_i | i=1, \cdots, N\}$, which is expressed as follows:
\begin{equation}
    \argmin_{\{\bm{\alpha_i}\}} \frac{1}{2}\|\bm{y}-\sum_{i=1}^N \bm{d}_i \|_2 + \lambda \sum_i \|\bm{\alpha}_i \|_1,
\end{equation}
where $\{\bm{\alpha_i}\}$ means the set of sparse maps, and the size of $\bm{\alpha}_i$ is the same as that of the image $\bm{y}$. To solve this optimization problem, many efficient algorithms~\cite{bristow2013fast, wohlberg2015efficient, wang2018scalable} have been proposed. The convolutional sparse coding model has been successfully adopted in many image processing tasks, such as image super-resolution~\cite{7410569}, image denoising~\cite{zheng2021deep} and multi-modal image restoration~\cite{xu2021deep}.

\citet{marivani2020multimodal} propose a multi-modal convolutional sparse representation model where each image modality is represented with its own dictionary filter, and different modalities are restricted to have similar sparse codes. Unlike \cite{marivani2020multimodal}, where each image modality is represented by a single dictionary, \citet{CUNet} propose to represent each image modality with two convolution dictionaries, where one dictionary is used to represent common features and the other is used to represent unique features. 
% The relationships between different modalities are as follows:
% \begin{equation}
%     \bm{y} = \sum_k\bm{d}_k^c * \bm{c}_k + \sum_k \bm{d}_k^u * \bm{u}_k, \quad \bm{g} = \sum_k\bm{h}_k^c * \bm{c}_k + \sum_k \bm{h}_k^u * \bm{v}_k,
% \end{equation}
% where $\{\bm{d}_k^c\}_{k=1}^K$ and $\{\bm{h}_k^c\}_{k=1}^K$ denote the common filters for the input depth map $\bm{y}$ and the guidance image $\bm{g}$, respectively; $*$ is the convolution operation; $\{\bm{d}_k^u\}_{k=1}^K$ and $\{\bm{h}_k^u\}_{k=1}^K$ denote the unique filters of $\bm{y}$ and $\bm{g}$, respectively. The common features share the same spare representations $\{\bm{c}_k\}_{k=1}^K$, while the unique features have their own spare representations $\{\bm{u}_k\}_{k=1}^K$ and $\{\bm{v}_k\}_{k=1}^K$. With the learned spare representations, the desired HR depth map $\bm{y}^*$ can be reconstructed as follows:
% \begin{equation}
%     \bm{y}^* = \sum_k \bm{g}_k^c * \bm{c}_k + \sum_k \bm{g}_k^u * \bm{u}_k.
% \end{equation}
The work of~\cite{CUNet} restricts the number of common and unique features to be the same; however, this may not always be true. To overcome this limitation, \citet{9506455} propose a model named CUNet++, which assigns different filters for different modalities. Moreover, they employ a recurrent architecture to extract the common and unique features to fully mimic the sparse model. \citet{9681224} propose an novel multi-modal convolutional dictionary learning model (MCDL), which models the correlation of different modalities at the image level. Similar to~\cite{CUNet}, they employ the common and unique dictionaries to represent each image modality. The difference is that the authors of \cite{9681224} propose to solve this multi-modal convolutional dictionary learning problem through traditional optimization, as they believe that the traditional optimization is more robust than the unfolding strategy when training data are limited.

\subsection{Deep Learning Methods}
In this subsection, we introduce the deep learning based methods. For this category, a straightforward way is to employ a two-stream network to extract features from the depth and color images separately, and then the extracted features are fused to recover the degraded depth maps. However, this simple framework may not have the ability to produce the desired high-resolution depth maps. To this end, researchers have developed various additional modules into this simple two-stream architecture to improve the depth map reconstruction performance. To better understand these methods, we introduce them from different perspectives as follows. (1) multi-modal information fusion strategy, (2) multi-task learning, (3) prior knowledge guided network design, (4) novel network architecture, (5) loss function.

\textbf{Multi-modal information fusion strategy.} The depth map and its corresponding color image are geometric and  photometric descriptions of the same scene, and they have strong structural similarity. Therefore, designing effective fusion algorithms to fuse these two images plays a significant role in guided depth map super-resolution. There are two widely used fusion tactics: single-level and multi-level fusion. \citet{GraphSR} fuse the multi-modal information at the date level, in which the original guidance and the pre-upscaled depth images are concatenated and sent to a convolution neural network for feature extraction. \citet{DKN} employ two different neural networks to learn the kernel weight and offset from the guidance image and the depth map, respectively. Then the learned kernel weight and offset are fused to generate the final weight and offset via element-wise multiplication. Instead of fusing multi-modal information at the image level, \citet{DJF, DJFR} propose to fuse multi-modal information at the feature level. Specifically, they first send the depth and color images to a two-stream network for feature extraction, then the extracted deep features are concatenated and sent to another network to generate the final result.

However, such a single-level fusion may not fully exploit the latent correlations between the depth and the guidance images. To address this problem, \citet{DMSG} propose a multi-scale fusion strategy, where the multi-scale guided features are obtained by a VGG-like neural network. Motivated by this, various works have been developed and have achieved promising performance. For example, ~\citet{DepthSR} propose to use a UNet based framework. In this framework, the encoder sub-network is used to extract features from the depth maps, while the decoder sub-network is used to fuse the guidance features at the corresponding scale. \citet{MIG} propose a multi-scale alternatively guided network, where the depth map is iteratively upsampled by $2\times$. In each upsampling, the depth features are alternatively guided by its corresponding gradient features and guidance features. \citet{zuo2019depth} propose a novel framework to improve the effectiveness of guidance information for the deeper network. First, they extract the multi-scale guidance features in the high-resolution to enhance the quality of guidance features. Moreover, they utilize dense connection to maintain multi-scale guidance features, thus, the depth features at each scale can be refined by guidance features from the current and all coarse scales.

The methods mentioned above mainly fuse the multi-modal features by concatenation or multiplication, which tends to produce texture-copying artifacts as they treat the features from different modalities equally. To solve this problem, ~\citet{AHMF} propose a multi-modal attention based fusion module (MMAF) to select and fuse the multi-modal features in a learning manner. The MMAF consists of a feature enhancement block and a feature recalibration block. The former is used to adaptively select informative feature from the guidance image, while the latter is designed to re-scale the multi-modal features so that the heterogeneous features can be directly fused by naive operations. \citet{song2020channel} propose an attention-based iterative residual learning framework to deal with the real-world depth map super-resolution. This framework contains an iterative residual learning module to progressively learn the high-frequency component for the depth map and an channel attention strategy to let the network pay more attention to the informative features. \citet{liu2021deformable} introduce an adaptive fully feature fusion module to enhance multi-scale feature representation ability. To fully exploit the complementary information contained in the depth and color image (\textit{i.e.}, enhance the edge information contained in the depth image and suppress the texture information in the color image), \citet{yang2022codon} propose a cross-domain attention
conciliation module (CAC) where the core is a parallel channel-spatial attention mechanism. \citet{li2020depth} propose a correlation-controlled color guidance block (CCGB) to improve the accuracy of the guidance image and avoid texture-transfer and depth-bleeding artifacts. The CCGB is made up of three major parts: correlation computing, weight learning, and guidance channel re-weighting.

\textbf{Multi-task learning.} Multi-task learning (MTL), also known as learning with auxiliary task, is a sub-filed of machine learning which aims to improve the performance of a specific task by simultaneously training multiple tasks. Motivated by this, some researchers propose to use the multi-task learning framework to boost the performance of depth map super-resolution. For example, \citet{tang2021bridgenet} propose a joint learning framework in which the tasks of depth map super-resolution (DSR) and depth estimation (DE) are solved at the same time. To make these two tasks cooperate with each other, the authors propose two different guidance strategies. One is the high-frequency attention bridge, which aims to learn the high-frequency information of the depth estimation task to guide the depth super-resolution process. The other is the content guidance bridge, which is designed to learn the content guidance from the depth super-resolution task to guide the depth estimation task. In their framework, these two tasks are trained separately with their corresponding loss functions. In a practical scenario, the paired RGB-D data may not always exist, especially for the aligned image pairs. In addition, processing the high-resolution guidance image will cause additional resource consumption. To solve these problems, \citet{sun2021learning} propose a novel cross-task distillation framework in which the guidance image is only required in the training stage. Similarly to~\cite{tang2021bridgenet}, they also employ depth estimation as an auxiliary task. Based on this, they further propose a cross-task interaction module to encourage the DE and DSR tasks to learn from each other. Unlike ~\cite{tang2021bridgenet}, the authors propose to use knowledge distillation to realize bilateral knowledge transfer between these two tasks. In this framework, the roles of teacher and student are exchanged between these tasks based on their current performance for depth map recovery, so that these two tasks can be tightly coupled and cooperated. When the model is trained, we can restore the low-resolution depth map with the DSR branch. 

Besides depth estimation, ~\citet{yan2022learning} propose to use depth completion as an auxiliary task to boost the performance of DSR. To make full use of the complementary information flows between these two tasks, they further design a novel joint correlation learning module (JCL) and an iterative-cross module (IC). The JCL is designed to capture long-range contextual dependencies and it is made up of a lightweight pixel-wise correlation and a channel-wise correlation. The IC is an iterative module which aims to reuse complementary information flows between the output layers.

\textbf{Prior knowledge guided network design.} Over the past few years, deep learning based depth map super-resolution methods have achieved unprecedented success due to their powerful representation ability. Despite promising performance, the black box nature makes it difficult to interpret, and performance improvements usually come from stacking new modules at the expense of increasing the model complexity. To alleviate this problem, some works propose to integrate the deep models into traditional optimization algorithms. For example, \citet{riegler2016deep} combine the advantages of the deep convolution neural network and the variational method and propose a deep primal-dual network. In their framework, they first use a stack of convolution layers to generate a guidance and a roughly super-resolved depth map, then the obtained guidance and depth images are sent to a non-local variational model to reconstruct the final depth map. \citet{zhou2023memory} propose a maximal posterior estimation model for GDSR with a local implicit prior and a global implicit prior. The local implicit prior is designed for modeling the complex relationship between the reconstructed depth image and the high-resolution guidance image from a local perspective, while the global implicit prior models the relationship between these two images from a global perspective. Moreover, they propose a persistent memory mechanism to improve the information
representation ability in both image and feature spaces.

\citet{marivani2020multimodal} propose to use convolution sparse representation to represent different image modalities at the image level. In their model, different modalities have different convolution dictionaries but with the same sparse representation. \citet{CUNet} propose to model each image modality with two convolution dictionaries, where the one is used to represent the common features and the other is used to model the unique features. \citet{GraphSR} propose to use a deep neural network to learn the affinity graph, and the learned graph is implemented as a differentiable optimization layer which can be used as a regularizer for depth map upsampling. This method can inherit the merits of both optimization and deep learning based methods: the optimization layer ensures the fidelity, and the deep neural network makes the learned graph have a large context. \citet{DCTNet} propose to use a DCT to solve a optimization model for guided depth map super-resolution, and then extend the solution of this optimization model to a deep neural network to solve the unexplainable problem of existing empirical-designed neural network. In addition, they propose a shared/private feature extraction module to extract cross-modal features and a guided edge-attention module to alleviate texture over-transfer problem.

\textbf{Novel network architecture.} Designing effective and efficient neural network architecture is a hot topic in deep learning, and in GDSR, researchers have tried numerous novel architectures to better exploit the intrinsic characteristics and correlations of multi-modal features. For example, \citet{zhao2019simultaneous} propose a color-depth conditional generative network (CDcGAN) to simultaneously super-resolve the low-resolution depth and color images. To reduce the artifacts caused by the differences of distributions between the depth and its corresponding color image, \citet{CGN} propose a depth-guided affine transformation network in which depth-guided color feature filtering and color-guided depth feature refinement are performed iteratively to progressively enhance the network representation ability. Moreover, all the refined depth features are concatenated to make full use of the iterations. Based on deep edge-aware learning, \citet{DSRN} propose a novel GDSR framework to preserve the sharpness of the depth edges. Their model first learns the edge information of the depth map, then, uses the learned edges to guide the depth reconstruction process. \citet{PMBAN} propose a progressive multi-branch aggregation network that consists of a reconstruction branch, a multi-scale branch, and a guidance branch. The reconstruction branch is an attention-based error feed-forward/-back module, which is used to iteratively highlight the informative features. The multi-scale branch aims to learn multi-scale representation. The outputs of these three branches are sent to a fusion block, which is actually a channel attention mechanism to adaptively select and fuse the valuable features. 

\citet{AHMF} propose to use a bi-directional RNN to encourage the cooperation multi-level features. \citet{tang2021bridgenet} propose a high-frequency attention bridge module and a content guidance bridge module to associate the tasks of depth super-resolution and depth estimation. \citet{JIIF} introduce the implicit image function into GDSR to improve the model capability. \citet{liu2021deformable} propose to use the deformable convolution~\cite{dai2017deformable} to align the depth features according to its corresponding guidance features. \citet{FDSR} propose a fast depth super-resolution baseline, where they use the octave convolution to fuse multi-modal features and reduce the computation complexity.~\citet{chen2019drop} propose to decompose the high-frequency component from the guidance features to guide the depth map reconstruction. \citet{li2020depth} propose a correlation-controlled color guidance block, which aims to improve the accuracy of the guidance information. \citet{yang2022codon} propose a cross-domain orchestration network (CODON) to effectively exploit the complementary information between depth and guidance features. Their model has two essential modules: recursive multi-scale convolutional residual module (RMC) and cross-domain attention conciliation module (CAC). The RMC is used to address the scale variance of scene structures, while the CAC is designed to effectively model co-occurrence correlations between depth and guidance features.

Although the above CNN-based methods have achieved significant success, there are still some limitations. First, existing CNN-based methods extract multi-modal features by applying cascaded convolutional layers on the input images. The local receptive field of convolutions makes it difficult for these methods to capture long-range spatial dependencies. Second, the weights of these convolutional layers are fixed during the inference process, which may hinder these methods from modeling the dynamic relationships between depth and guidance images. To mitigate the limitations mentioned above of CNN-based methods, some Transformer-based approaches are proposed in the area of GDSR. For example, \citet{9428393} propose a texture-depth transformer to explore useful texture information for depth super-resolution. Specifically, the authors first use a pretrained VGG model to extract multi-scale features. Then, the extract features are sent to the proposed texture-depth transformer module to transfer the structures from the guidance features to the depth features. Finally, these enhanced features are used to generate the desired depth map. For depth map super-resolution, it is vital to maintain high-resolution representations. However, the computational complexity of self-attention is quadratic to the size of the image. Directly applying the Transformer to GDSR task would result in a great waste of computational resources. To overcome this issue, \citet{ariav2022depth} propose a cascaded transformer-based module for GDSR where they calculate self-attention in a local window and capture long-range dependencies with a shift operation. The complexity of this window-based attention is linear. Thus, their model is capable of handling large inputs.

\textbf{Loss function}. In the deep learning based depth super-resolution methods, the most widely used loss functions are the pixel-wise $\mathcal{L}_1$ and $\mathcal{L}_2$ losses, since they result in lower MAE/RMSE values. The $\mathcal{L}_1$ and $\mathcal{L}_2$ losses are defined as follows:
\begin{equation}
     \mathcal{L}_1 = \frac{1}{N} \sum_{i=1}^N \lvert \bm{X}_{i}-\bm{Y}^{*}_i \rvert, \quad \mathcal{L}_2 = \frac{1}{N} \sum_{i=1}^N (\bm{X}_{i}-\bm{Y}^{*}_i)^2,
\end{equation}
where $N$ is the number of pixels, $\bm{X}$ means the ground-truth and $\bm{Y}^{*}$ denotes the reconstructed depth map. However, since these pixel-wise losses treat every pixel equally, models trained with these functions tend to generate over-smooth results. Motivated by this, various loss functions are proposed to force their model to generate more realistic results. For example, \citet{sun2021learning} propose to use SSIM loss to improve the structure similarity between the reconstructed depth map and the ground-truth. The SSIM loss is as follows:
\begin{equation}
    \mathcal{L}_{\mathrm{SSIM}} = 1 - \mathrm{SSIM}(\bm{X}, \bm{Y}^{*}), 
\end{equation}
where SSIM means the SSIM metric, and it is defined in Eq.~\ref{ssim}. \citet{zhao2019simultaneous} propose a 8-neighboring gradient difference (GD)
loss to force the network to pay more attention to the boundaries, and the GD loss is formulated as follows:
\begin{equation}
    \mathcal{L}_{\mathrm{GD}} = \frac{1}{N} \sum_i(\sum_{k\in \Omega} |\nabla_k \bm{X}_i - \nabla_k \bm{Y}^{*}_i|_1), 
\end{equation}
where $\Omega$ means the pixel's neighbourhood, $\nabla_k$ denotes the $k$-th gradient between each pixel and $k$-th pixels among 8-neighbouring pixels. \citet{dong2022learning} propose a cycle-consistent loss for self-supervised depth map restoration:
\begin{equation}
    \mathcal{L}_{\mathrm{C}} = |f_{\mathrm{down}}(f_{\mathrm{net}}(\hat{\bm{Y}}, \bm{G})) - \hat{\bm{Y}}|_1,
\end{equation}
where $\hat{\bm{Y}}$ means the interpolated low-resolution depth map, $f_{\mathrm{down}}$ is average pooling downsampling and $f_{\mathrm{net}}$ denotes the reconstruction network.

\section{Depth related applications}
\label{application}
In the following, we introduce depth-related application briefly, including salient object detection, semantic segmentation and \textit{etc}.

\subsection{Salient Object Detection}
 
RGB image contains rich color and content information while depth image can provide supplementary depth information. In light of this, many works explore the potential of depth cue in salient object detection. To explore depth cue and inherit the performance RGB salient object detection, \citet{8807367} develop a transformation model which helps existing RGB model work well on RGB-D scenarios. Considering that RGB and depth images belong to two different modalities, \citet{10.1145/3474085.3475364} build RGB-induced and depth-induced Enhancement modules to differentially models the dependence of two modalities. \citet{9930882} couple interaction and refinement to explore cross-modalities information while \citet{9686679} perform selient object detection in a bi-directional progressive manner. \citet{9810116} propose to extract intra- and inter-saliency cues which are used to promote salient object detection.
 
\subsection{Semantic Segmentation} Depth image contains geometric information which is important for semantic segmentation, depth image has been introduced into semantic segmentation. \citet{10.1007/978-3-030-58621-8_33} propose a separation-and-aggregation gate to fuse multi-modal information and utilize bi-direction multi-step propagation to promote information interaction. \citet{LI2021209} design a collaborative optimization method to generate segmentation results. \citet{ZHOU2022108468} develop two encoders to extract information from RGB and depth images and carefully design co-attention mechanism to fuse color and depth information. \citet{9345730} propose to utilize depth and RGB images to generate 2D semantic segmentation results and then obtain refined 3D semantic segmentation results with 2D ones.

\section{Experiments}\label{comparison}

In this section, we conduct experiments to evaluate the performance of representative guided depth map super-resolution (GDSR) approaches. Firstly, we introduce the datasets and evaluation metrics used in this paper. Besides, both the quantitative and qualitative results for the compared methods are presented. Based on these results, we analysis the advantages and disadvantages between different types of method. Finally, we conducts experiments with several tricks widely used in other vision tasks and explore whether they can improve the performance of GDSR.

\textbf{Datasets and Metrics}. Three widely used datasets are selected, including Middlebury dataset~\cite{2001, 2003, 2005, 2006, 2014}, NYU v2 dataset~\cite{NYU} and RGB-D-D dataset~\cite{FDSR}. For Middlebury dataset, following the experimental settings of~\cite{DMSG, DepthSR}, we utilize 82 RGB-D image pairs collected by~\cite{DMSG} as the training set, and 6 RGB-D image pairs from Middlebury 2005~\cite{2005} as the testing set. However, this dataset only contains six RGB-D image pairs for evaluation, which cannot fully demonstrated the performance of deep learning based methods. To this end, we further conduct experiments on a large scale NYU v2 dataset~\cite{NYU} to compare recently proposed deep learning based methods. Specifically, following~\cite{DJFR, DKN}, we use the first 1000 RGB-D image pairs from the this dataset as the training data and the testing data are: 1) the rest of 449 image pairs from the NYU v2 dataset~\cite{NYU}; 2) 1064 image pairs from the Sintel dataset~\cite{butler2012naturalistic}; 3) 325 image pairs from the DIDOE indoor test set~\cite{DIDOE}; 4) the first of 500 image pairs from the SUN RGBD test set~\cite{SUN}; 5) 405 image pairs from the RGB-D-D test set~\cite{FDSR}, 6) 503 image pairs from the DIML indoor test set~\cite{cho2021deep}. In addition, we also conduct experiments on RGB-D-D dataset~\cite{FDSR}, and this is a newly proposed real-world dataset.

Among these dataset, the Middlebury and the NYUv2 datasets are synthetic dataset, we use Bicubic interpolation to generate the LR depth maps, and the downsampling factors are set as $4\times, 8\times$ and $16\times$. For the RGB-D-D dataset, it is a real-world dataset that contains paired LR-HR depth maps, and we use the official split training/testing data to train and evaluate the GDSR models. For both datasets, we follow existing state-of-the-art methods~\cite{DJFR, DMSG, 8491336, SVLRM, FDSR} that employ RMSE values as the image quality assessment metric. 

% It should be noted that all the compared methods are trained and evaluated on the same datasets, and all the deep learning-based methods are trained with $\mathcal{L}_1$ loss and optimized with Adam optimizer for fair comparison. We use the RMSE values as the image quality assessment metric.

\begin{table}[!t]\setlength{\tabcolsep}{2.5pt}\renewcommand{\arraystretch}{1.1}\renewcommand\theadfont{\footnotesize}
    \footnotesize 
    \centering
    \caption{Quantitative comparison with the state-of-the-art methods on Middlebury datasets~\cite{2005}. We use RMSE metric (The lower the better). The best performance is shown in \textbf{bold}, while the second and the third best performance are the \underline{underscored} and the \uwave{waved} ones, respectively.}
    \vspace{-0.1in}
    \begin{tabular}{lcccccccccccccccccc}
		\toprule
		\multirow{2}{*}{Method} & \multicolumn{3}{c}{Art} & \multicolumn{3}{c}{Book} & \multicolumn{3}{c}{Dools}  & \multicolumn{3}{c}{Laundry} & \multicolumn{3}{c}{Mobeius}& \multicolumn{3}{c}{Reindeer}\\
		\cmidrule(lr){2-4} \cmidrule(lr){5-7} \cmidrule(lr){8-10} \cmidrule(lr){11-13} \cmidrule(lr){14-16} \cmidrule(lr){17-19} 
		~ &$4\times$ & $8 \times$  & $16\times$  & $4\times$ & $8 \times$  & $16\times$  &$4\times$ & $8 \times$  & $16\times$ &$4 \times$ & $8 \times$  & $16\times$ &$4 \times$ & $8 \times$  & $16\times$ &$4 \times$ & $8 \times$  & $16\times$ \\
		\hline
        Bicubic & 3.87 & 5.46 & 8.17 & 1.27 & 2.34 & 3.34 & 1.31 & 1.86 & 2.62 & 2.06 & 3.45 & 5.07 & 1.33 & 1.97 & 2.85 & 2.42 & 3.99 & 5.86 \\
        GF~\cite{GF} &  3.78 &5.41 &8.04 &1.29 &2.26 &3.34 &1.30 &1.85 &2.61 &2.01 &3.40 &5.01 &1.32 & 1.83 & 2.79 & 2.32 & 3.88 &5.86  \\
        SDF~\cite{zhang2015segment} & 3.55 &5.30 &7.84 &1.24 &2.31 &3.33 &1.29 &1.84 &2.61 &1.91 &3.37 &5.04 &1.27 &1.91 &2.83 &2.33 &3.93 &5.84 \\
        TGV~\cite{riegler2016atgv} & 3.34 & 4.10 & 6.43 & 1.47 & 1.82 & 2.63 & 1.31 & 1.61 & 2.22 & 2.39 & 2.64 & 4.17 & 1.22 & 1.64 & 2.41 & 2.71 & 3.15 & 4.60  \\
        AR~\cite{6827958} & 2.80 & 3.65 & 5.90 & 1.54 & 1.82 & 2.92 & 1.38 & 1.66 & 2.34 & 2.22 & 2.69 & 4.57  & 1.01 & 1.46 & 2.53 & 2.45 & 2.96 & 4.11\\
        SRAM~\cite{wang2019multi}& 2.57 & 3.20 & 4.87 & 1.33 & 1.46 & 2.51 &  1.07 & 1.19 & 1.90 & 2.00 & 2.11 & 4.07 & 0.85 & 1.10 & 1.98 & 2.07 & 2.47 & 3.44  \\
        JGIE~\cite{8491336} & 2.02 & 3.38 & 5.58 & 0.93 & 1.57 & 2.47 & 0.97 & 1.27 & 1.76 & 1.33& 2.00 & 2.37 & 0.97 & 1.42 & 2.19 & 1.39 & 2.23 & 3.56 \\
        SRCNN~\cite{SRCNN} & 1.87 & 3.70 & 7.31 & 0.85 & 1.59 & 3.12 & 0.88 & 1.46 & 2.42 &1.08 & 2.31 & 4.60 & 0.86 & 1.48 & 2.68 & 1.35 & 2.74 & 5.33 \\
        DJFR~\cite{DJFR} &1.62 & 3.08 & 5.81 & 0.54 & 1.11 & 2.24 & 0.78 & 1.27 & 2.02 & 0.90 & 1.83 & 3.65 & 0.68 & 1.22 & 2.21 & 1.25 & 2.38 & 4.22   \\ 
        DMSG~\cite{DMSG} & 1.47 & 2.46 & 4.57 & 0.67 & 1.03 & 1.60 & \underline{0.69} & \underline{1.05} & \underline{1.60} & \underline{0.79} & 1.51 & 2.63 & 0.66 & 1.02 & 1.63 & 0.98 & 1.76 & 2.92 \\
        DepthSR~\cite{DepthSR} & 1.20 &2.22 &  \textbf{3.90} & 0.60 & 0.89 & \underline{1.51} & 0.84 & 1.14 & \textbf{1.52} & 0.78 & \underline{1.31} & 2.26 & 0.96 & 1.19 & \underline{1.58} & 0.96 & \underline{1.57} & \textbf{2.47}   \\
        CGN~\cite{CGN} & 1.50 & 2.69 & \underline{4.14} & 0.60 & 0.97 & 1.73 & 0.88 & 1.20 & 1.80 & 0.98 & 1.57 & 2.57 & 0.69 & 1.06 & 1.69 & 1.22 & 2.02 & 3.60   \\
        MFR-SR~\cite{8598786} & 1.54 & 2.71 & 4.35 & 0.63 & 1.05 & 1.78 & 0.89 & 1.22 & 1.74 & 1.23 & 2.06 & 3.74 & 0.72 & 1.10 & 1.73 & 1.23 & 2.06 & 3.74 \\
        %DKN ~\cite{DKN} & \underline{1.12} & \underline{2.46} & 4.61 & 0.44 & \underline{0.82} & 1.71 & 0.71 & 1.14 & 1.75 & \underline{0.79} & 1.46 & \underline{2.23} & 0.59 & \underline{0.97} & 1.68 & \underline{0.92} & 1.83 & 3.30 \\
        AHMF~\cite{AHMF} & \textbf{1.09} & \textbf{2.14} & 4.20 & \textbf{0.38} & \textbf{0.72} & \textbf{1.49} & \textbf{0.62} & \textbf{1.03} & \underline{1.66} & \textbf{0.64} & \textbf{1.22} & \textbf{2.14} & \textbf{0.54} & \textbf{0.88} & \textbf{1.53} & \textbf{0.85} & \textbf{1.56} & \underline{2.84} \\
	\bottomrule
	\end{tabular}
    \label{md_rmse}
    \vspace{-0.1in}
\end{table}
\begin{figure}[!htb]
    \centering
    \includegraphics[width=\linewidth]{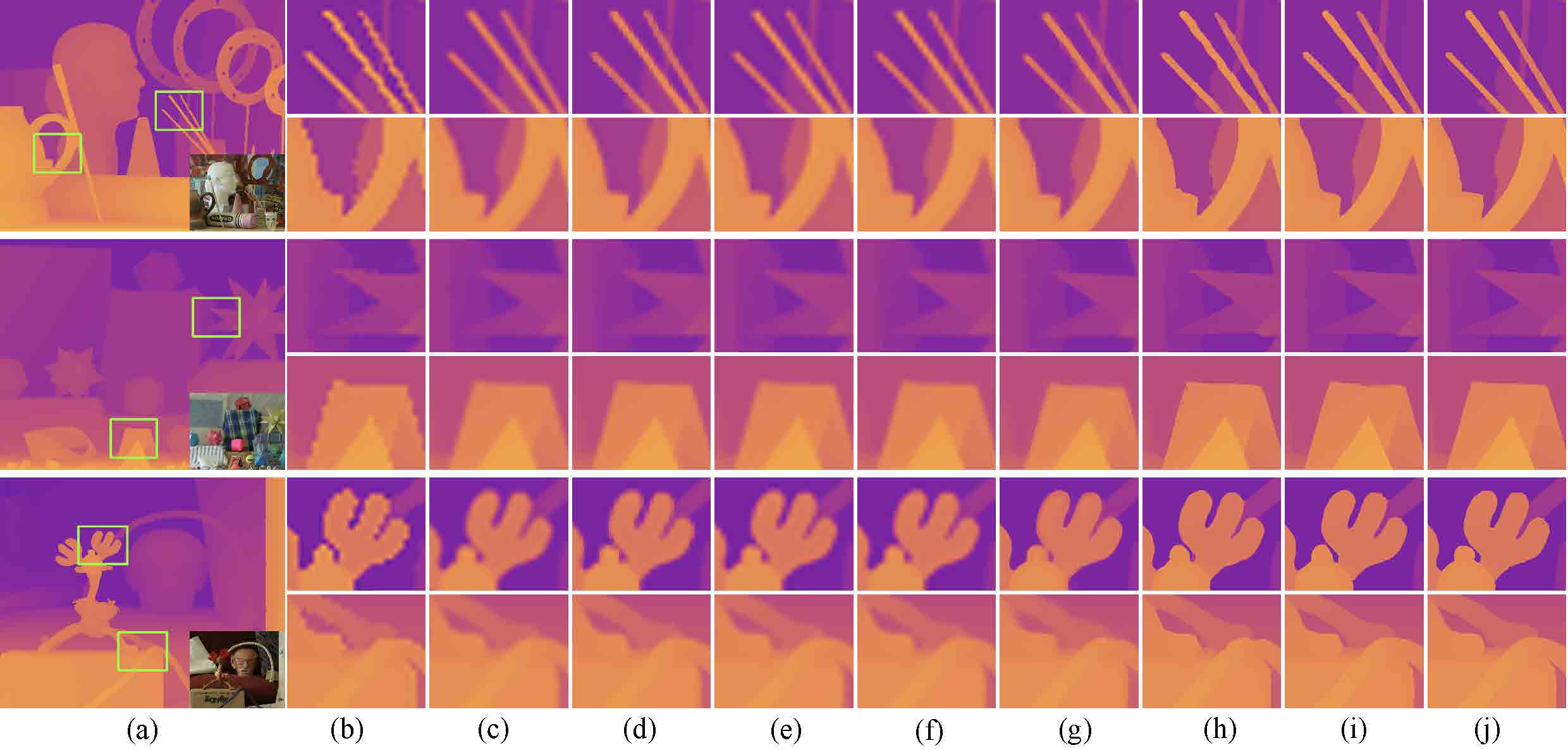}
    \vspace{-0.32in}
    \caption{Visual comparisons for $8\times$ guided depth map super-resolution on \textit{Art, Moebius} and \textit{Reindeer} from Middlebury dataset. (a) GT and RGB images, (b) Bicubic, (c) GF~\cite{GF}, (d) SDF~\cite{zhang2015segment}, (e) TGV~\cite{riegler2016atgv}, (f) JGIE~\cite{8491336}, (g) DJFR~\cite{DJFR}, (h) AHMF~\cite{AHMF}, (i) GT. Please enlarge the PDF for more details.}
    \label{fig:mb}
    \vspace{-0.1in}
\end{figure}
\begin{table}[!t]\setlength{\tabcolsep}{2.2pt}\renewcommand{\arraystretch}{1.1}\renewcommand\theadfont{\footnotesize}
    \footnotesize 
    \centering
    \caption{Quantitative comparison with the state-of-the-art methods on NYU v2~\cite{NYU}, MPI Sintel Depth~\cite{butler2012naturalistic}, DIDOE~\cite{DIDOE}, SUN RGB-D~\cite{SUN}, RGB-D-D~\cite{FDSR} and DIML~\cite{cho2021deep} datasets for $\times 4, \times 8$ and $\times 16$ scale factors, respectively. We use RMSE metric (The lower the better). The best performance is shown in \textbf{bold}, while the second and the third best performance are the \underline{underscored} and the \uwave{waved} ones, respectively.}
    \vspace{-0.1in}
    \begin{tabular}{lcccccccccccccccccc}
		\toprule
		\multirow{2}{*}{Method} & \multicolumn{3}{c}{NYU v2} & \multicolumn{3}{c}{Sintel} & \multicolumn{3}{c}{DIDOE}  & \multicolumn{3}{c}{SUN RGB} & \multicolumn{3}{c}{RGB-D-D}& \multicolumn{3}{c}{DIML}\\
		\cmidrule(lr){2-4} \cmidrule(lr){5-7} \cmidrule(lr){8-10} \cmidrule(lr){11-13} \cmidrule(lr){14-16} \cmidrule(lr){17-19} 
		~ &$4\times$ & $8 \times$  & $16\times$  & $4\times$ & $8 \times$  & $16\times$  &$4\times$ & $8 \times$  & $16\times$ &$4 \times$ & $8 \times$  & $16\times$ &$4 \times$ & $8 \times$  & $16\times$ &$4 \times$ & $8 \times$  & $16\times$ \\
		\hline
        Bicubic & 4.27 &7.17 &11.37 & 6.41 &8.79 &11.90 & 5.67 &8.41 &10.01 & 1.23 &2.01 &3.19 & 2.75 &4.47 &6.98 & 1.92 &3.20 &5.14 \\
        GF~\cite{GF}& 5.05 & 6.97 & 11.13  & 7.15 & 8.66 & 11.63 & 7.80 & 8.78 & 9.98 & 1.75 & 2.18 & 3.21 & 2.72 & 4.02 & 6.68 & 2.72 & 3.40 & 5.13 \\
        % MuGIF~\cite{muGIF} \\
        DGF~\cite{DGF} & 3.21 & 5.92 & 10.45 & 5.91 & 8.02 & 11.17 & 5.79 & 8.41 & 9.94 & 1.16 & 1.95 & 3.14 & 2.11 & 3.56 & 6.31 & 2.01 & 3.22 & 5.10\\
        DMSG~\cite{DMSG} &  3.02 & 5.38 & 9.17 & 4.73& 6.26 & 8.36 & 6.05 & 8.47 & 9.83 & 0.91 & 1.53 & 2.62 & 1.80 & 3.04 & 5.10 & 1.39 & 2.34 & 4.02 \\
        DJF~\cite{DJF} & 2.80 & 5.33 & 9.46 & 5.30 & 7.53 & 10.41 & 6.04 & 8.29 & 10.15 & 0.89 & 1.65 & 2.83 & 1.72 & 2.96 & 5.26 & 1.39 & 2.49 & 4.27 \\
        DepthSR~\cite{DepthSR} &3.00 &  5.16 &  8.41 & 4.49 &  6.53 &  9.28 & 6.07 & 8.44 & 9.73 & 0.92 & 1.45 & 2.48 & 1.82 & 2.85 & 4.60 & 1.40 & 2.23 & 3.75 \\
        DJFR~\cite{DJFR}& 2.38 & 4.94 & 9.18 & 4.90 & 7.39 & 10.33 & 5.63 & 8.24 & 9.89 & 0.81 & 1.54 & 2.80 & 1.50 & 2.72 & 5.05 & 1.27 & 2.34 & 4.13 \\
        %PMBAN~\cite{PMBAN}& 1.20 & 2.64 & 4.94 & 3.52 & 5.10 & 7.29 & 2.71 & 5.31 & 7.96 & 0.52 & 0.92 & 1.69 & 1.06 & 1.67 & 3.09 & 1.08 & 1.63 & 2.52 \\
        CUNet~\cite{CUNet}& 1.65 & 3.35 & 6.64 & 4.27 & 5.96 & 8.51 & 3.80 & 6.98 & 9.44 & 0.63 & 1.13 & 2.20 & 1.20 & 1.86 & 3.27 & 1.18 & 1.88 & 3.25 \\
        SVLRM~\cite{SVLRM}& \uwave{1.51} & 3.21 & 6.98 & \uwave{4.05} & 5.83 & 8.60 & 3.58 & 6.96 & 9.55 & \uwave{0.59} & 1.10 & 2.33 & 1.22 & 1.88 & 3.55 & 1.19 & 1.93 & 3.49 \\
        FDKN~\cite{DKN} & 1.86 & 3.58 & 6.96 & 4.23 & 5.92 & 8.73 & 4.04 & 7.07 & 9.50 & 0.67 & 1.13 & 2.23 & 1.18 & 1.91 & 3.41 & \underline{1.13} & 1.84 & 3.29\\
        DKN~\cite{DKN}& 1.62 & 3.26 & 6.51 & 4.38 & 5.89 & 8.40 & \uwave{3.49} & 6.96 & 9.31 & 0.63 & 1.10 & 2.16 & 1.31 & 1.87 & 3.26 & 1.27 & 1.86 & 3.22 \\
        GraphSR~\cite{GraphSR} & 1.79 & \uwave{3.04} & 6.02 & 4.29 & \underline{5.56} & 7.93 & 3.93 & \uwave{6.37} & 9.06 & 0.67 & \uwave{1.05} & 2.03 & 1.30 & \uwave{1.83} & 3.12 & 1.25 & 1.79 & 3.03\\
        FDSR~\cite{FDSR} & 1.61 & 3.18 & \uwave{5.86} & 4.14 & 5.67 & \uwave{7.86} & 3.62 & 6.54 & \uwave{8.85} & 0.64 & \uwave{1.05} & \uwave{1.97} & \uwave{1.16} & 1.82 & \uwave{3.06} & \textbf{1.10} & \underline{1.71} & \uwave{2.87}\\
        % DCTNet~\cite{DCTNet}&1.59 & 3.08 & \uwave{5.80} & 4.18 & 6.31 & 9.16 & 3.84 & 7.20 & 9.69 & 0.65 & 1.28 & 2.43 & \textbf{1.07} & \uwave{1.78} & 3.18 & \textbf{1.07} & \underline{1.71} & 2.99  \\
        AHMF~\cite{AHMF}& \underline{1.40} & \underline{2.89} & \underline{5.64} & \underline{3.84} & \uwave{5.62} & \underline{7.55} & \textbf{2.93} & \textbf{6.14} & \textbf{8.54} & \underline{0.57} & \underline{0.99} & \underline{1.82} & \textbf{1.10} & \textbf{1.73} & \underline{3.04} & \textbf{1.10} & \textbf{1.70} & \textbf{2.83} \\
        JIIF~\cite{JIIF} & \textbf{1.37} & \textbf{2.76} & \textbf{5.27} & \textbf{3.82} & \textbf{5.50} & \textbf{7.46} & \underline{2.94} & \underline{6.17} & \underline{8.58} & \textbf{0.54} & \textbf{0.95} & \textbf{1.79} & \underline{1.15} & \underline{1.77} & \textbf{2.79} & \uwave{1.17} & \uwave{1.79} & \underline{2.86}\\
		\bottomrule
		\end{tabular}
    \label{nyu_rmse}
    \vspace{-0.1in}
\end{table}
\begin{figure}[!tb]
	\begin{center}
		\subfigure[~]{
		\begin{minipage}[b]{0.33\linewidth}
			\includegraphics[width=1\linewidth]{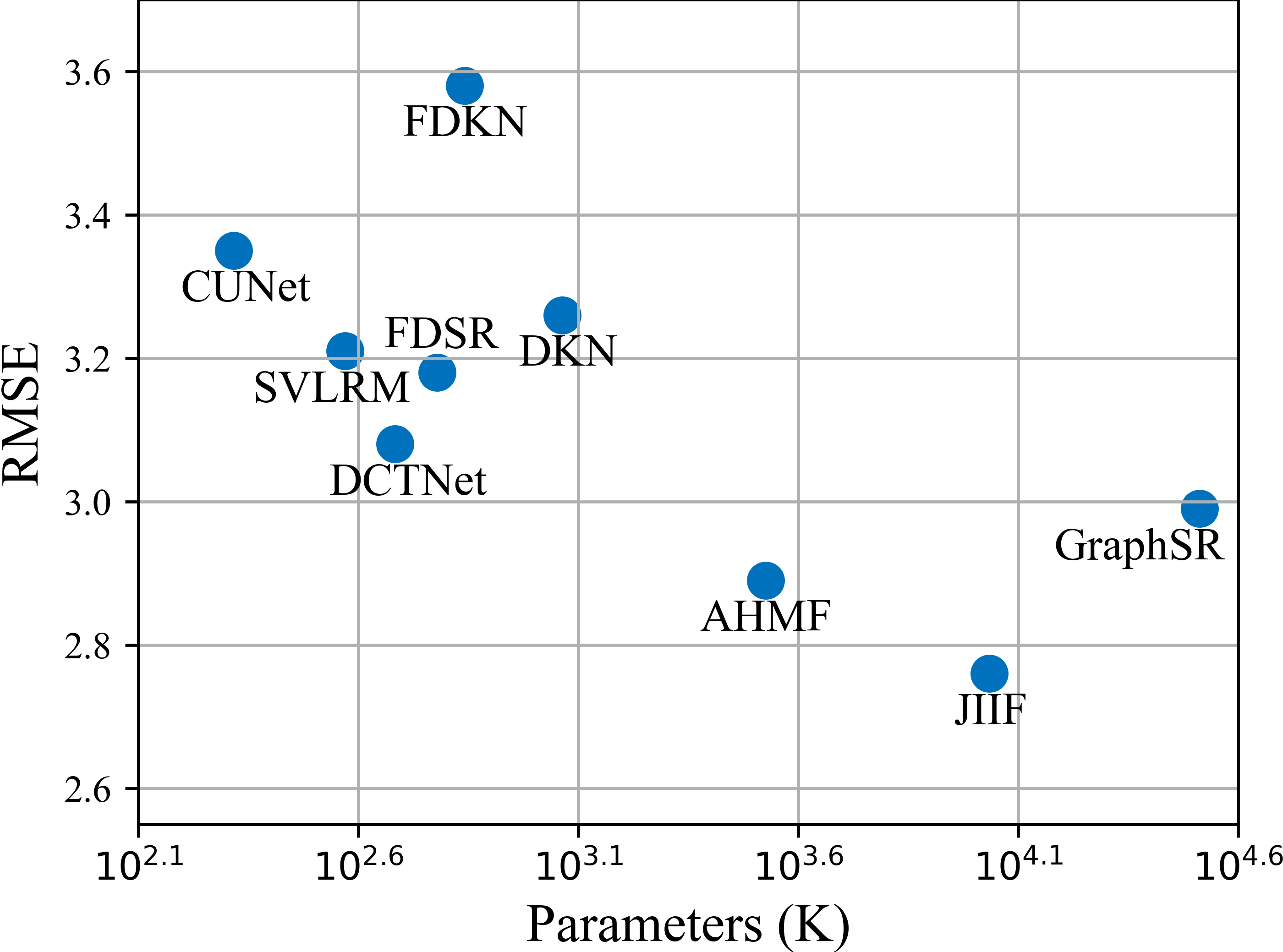} 
		\end{minipage}
		\hspace{-0.1in}
	}\subfigure[~]{
    		\begin{minipage}[b]{0.33\linewidth}
  		 	\includegraphics[width=1\linewidth]{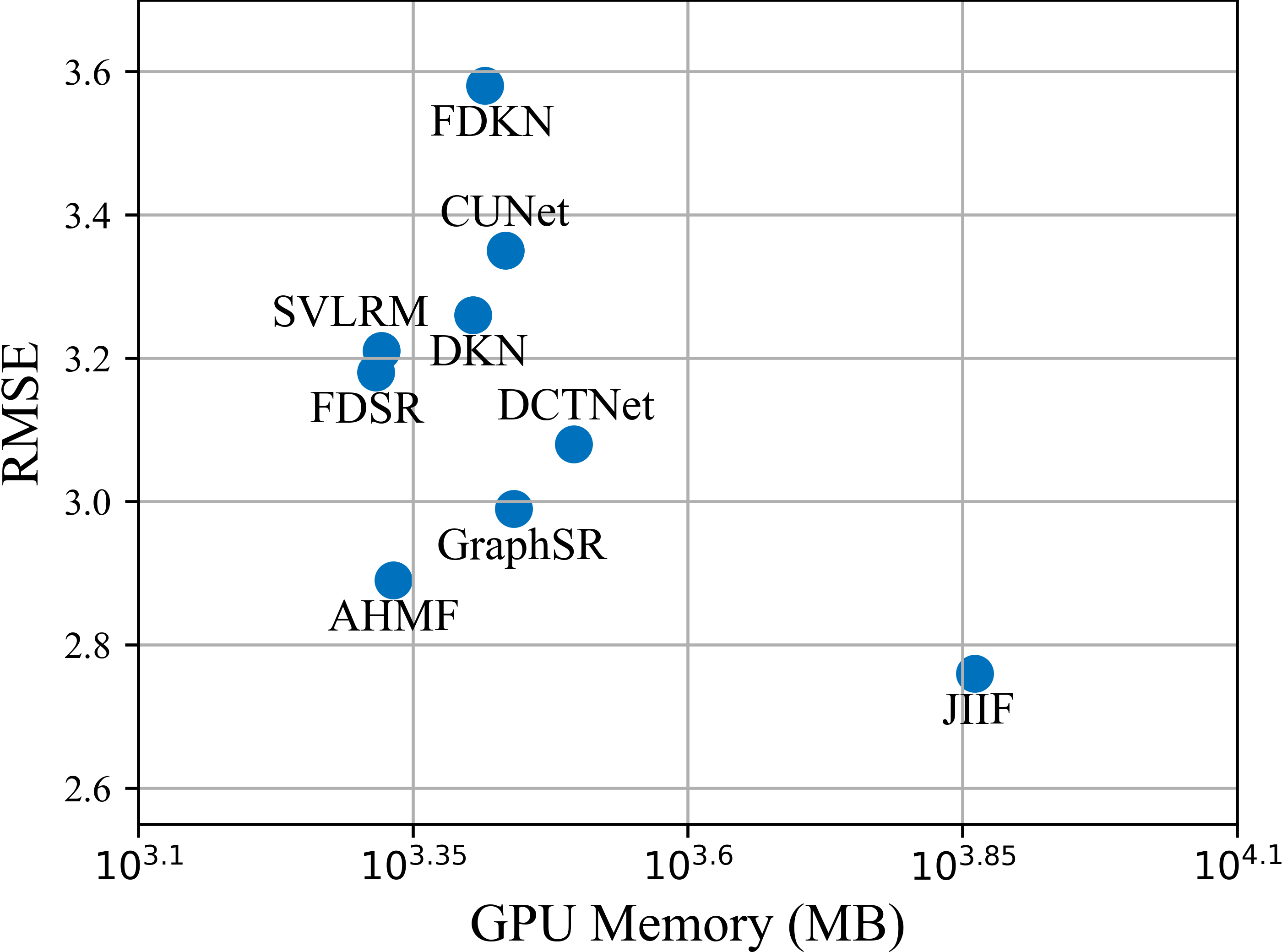}
    		\end{minipage}
    		\hspace{-0.1in}
    	}\subfigure[~]{
    		\begin{minipage}[b]{0.33\linewidth}
  		 	\includegraphics[width=1\linewidth]{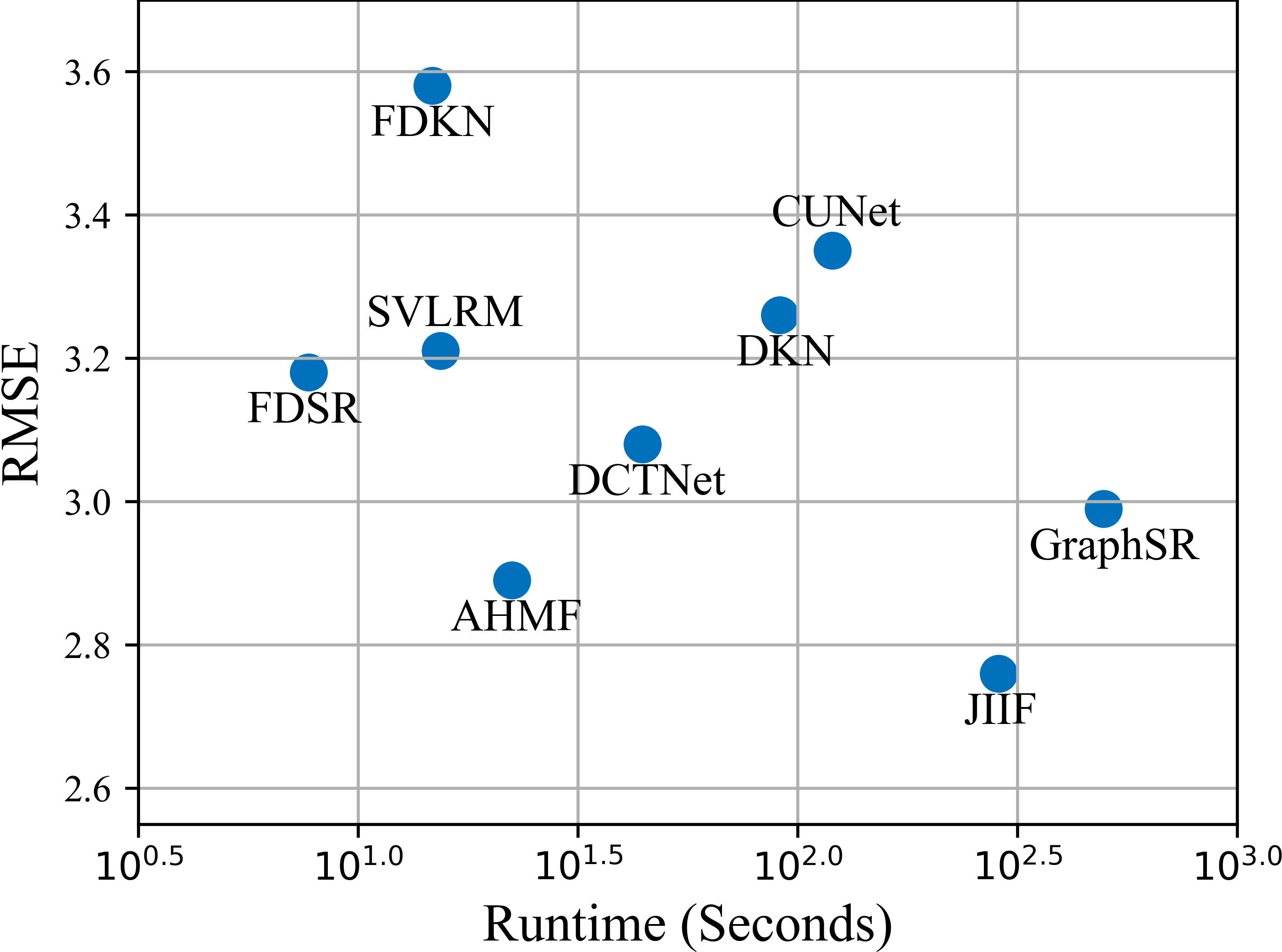}
    		\end{minipage}
    	}
	\end{center}
	\vspace{-.1in}
	\caption{Comparison of state-of-the-arts for $8\times$ GDSR on NYU v2~\cite{NYU} dataset in terms of RMSE, number of parameters (a), peak GPU memory consumption (b), and average running times (c). The experiments are evaluated on a NVIDIA 3090 GPU with HR depth map size $480 \times 640$.}
	\label{rmse_time}
	\vspace{-0.1in}
\end{figure}

\begin{table}[!t]\setlength{\tabcolsep}{2.4pt}\renewcommand{\arraystretch}{1.1}\renewcommand\theadfont{\footnotesize}
    \footnotesize 
    \centering
    \caption{Quantitative comparison with the state-of-the-art methods on RGB-D-D~\cite{FDSR} dataset. We use RMSE metric (The lower the better). The best performance is shown in \textbf{bold}, while the second and the third best performance are the \underline{underscored} and the \uwave{waved} ones, respectively.}
    \vspace{-0.1in}
    \begin{tabular}{lccccccccc}
		\toprule
        Method & Bicubic & GF~\cite{GF} & DJFR~\cite{DJFR} & SVLRM~\cite{SVLRM} & DKN~\cite{DKN} & FDSR~\cite{FDSR} & AHMF~\cite{AHMF} & GraphSR~\cite{GraphSR} & DCTNet~\cite{DCTNet} \\
        \midrule
        RMSE & 7.19 & 6.97 & 6.24 & 5.81  &5.74 & \uwave{5.49} & \underline{5.46} & 5.52 & \textbf{5.38} \\
		\bottomrule
		\end{tabular}
    \label{tab:rd_rmse}
    \vspace{-0.2in}
\end{table}

\textbf{Experimental Results}. We first show the experimental results on Middlebury dataset~\cite{2005}. For this dataset, we select 13 state-of-the-art methods as competitors, which includes two local filtering-based methods (\textit{i.e.}, GF~\cite{GF} and SDF~\cite{zhang2015segment}), four optimization-based methods (\textit{i.e.}, TGV~\cite{riegler2016atgv}, AR~\cite{6827958}, SRAM~\cite{wang2019multi} and JGIE~\cite{8491336}) and seven learning-based methods (\textit{i.e.}, SRCNN~\cite{SRCNN}, DJFR~\cite{DJFR}, DMSG~\cite{DMSG}, DepthSR~\cite{DepthSR}, CGN~\cite{CGN}, MFR-SR~\cite{8598786} and AHMF~\cite{AHMF}). We set Bicubic interpolation as the baseline model. The quantitative results for $4\times, 8\times$ and $16\times$ depth map super-resolution are presented in Tab~\ref{md_rmse}. It can be observed that the learning-based methods obtain the best performance due to the powerful learning ability of the deep convolution neural network. The optimization-based methods are more robust than the filtering based methods. The single depth map super-resolution method (\textit{i.e.}, SRCNN~\cite{2005}) has a complex network architecture, achieves better performance than some traditional methods for the small up-scale factor. However, when the up-scale factor increases, the low-resolution depth maps are seriously damaged, making it impossible to generate accurate depth maps. Thus, the performance of SRCNN~\cite{SRCNN} is worse than these of the compared methods. By introducing the guidance information, the performance of DJFR~\cite{DJFR} is significantly improved than SRCNN~\cite{SRCNN}. Compared to DJFR~\cite{DJFR}, the subsequent methods (\textit{e.g.}, DepthSR~\cite{DepthSR} and CGN~\cite{CGN}) adopt powerful network architectures and elaborately designed guidance strategies, thus the depth map reconstruction performance is further improved.

Fig~\ref{fig:mb} shows the upsampled depth maps ($8\times$) for \textit{Art}, \textit{Moebius}, and \textit{Reindeer}. The visual comparisons are generally consistent with the quantitative results that the learning-based methods achieve better performance than the conventional hand-crafted methods. The GF~\cite{GF} suffer from halo artifacts due to its local characteristic. The results of SDF~\cite{zhang2015segment} suffer from jagging artifacts at the pencial region. The results of TGV~\cite{riegler2016atgv} are too smooth, which means that the hand-designed regularizer cannot deal with large up-scale factor. The JGIE~\cite{8491336} combines both the local structure and non-local low-rank regularization, and obtains better performance than TGV~\cite{riegler2016atgv}, which implies that combining multiple regularizers is beneficial for GDSR. The results of DJFR~\cite{DJFR} are more faithful to the ground-truth image. However, they still suffer from texture copying artifacts in some regions, since the DJFR~\cite{DJFR} uses naive method to fuse the guidance information. The AHMF~\cite{AHMF} proposes an attention-based fuse mechanism to fuse the guidance information, achieving the best performance.

\begin{figure}[!tb]
	\begin{center}
		\subfigure[\scriptsize{Guidance}]{
		\begin{minipage}[b]{0.12\linewidth}
			\includegraphics[width=1\linewidth]{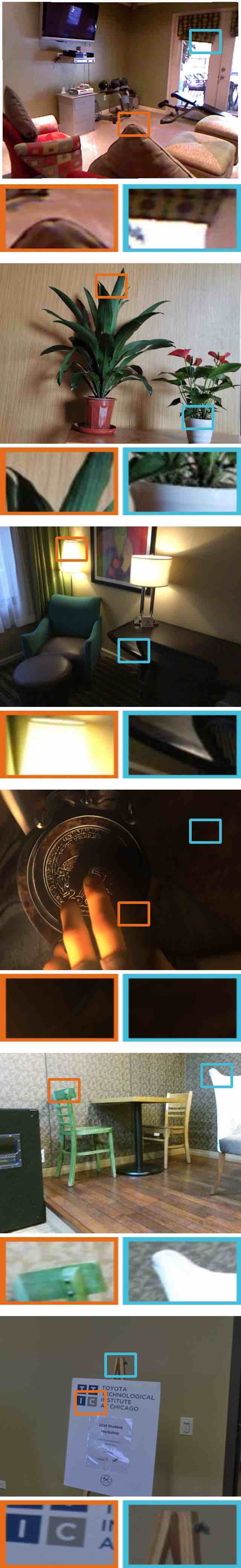} 
		\end{minipage}
		\hspace{-0.08in}
	    }\subfigure[\scriptsize{Bicubic}]{
    		\begin{minipage}[b]{0.12\linewidth}
  		 	\includegraphics[width=1\linewidth]{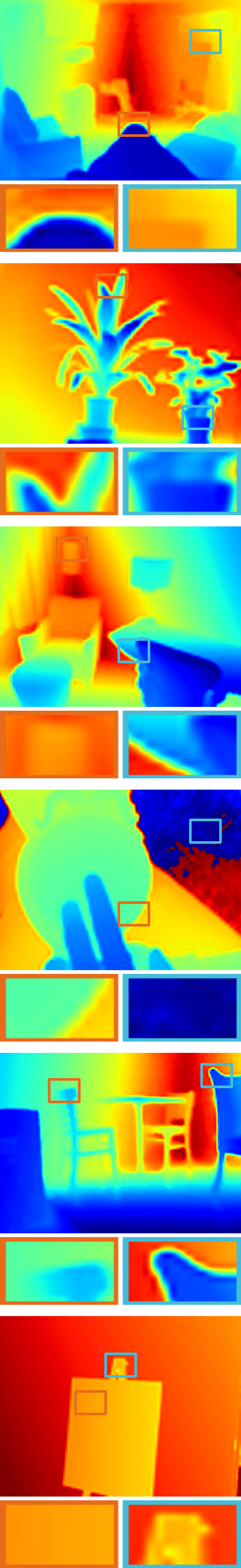}
    		\end{minipage}
    		\hspace{-0.08in}
    	}\subfigure[\scriptsize{GF~\cite{GF}}]{
    		\begin{minipage}[b]{0.12\linewidth}
  		 	\includegraphics[width=1\linewidth]{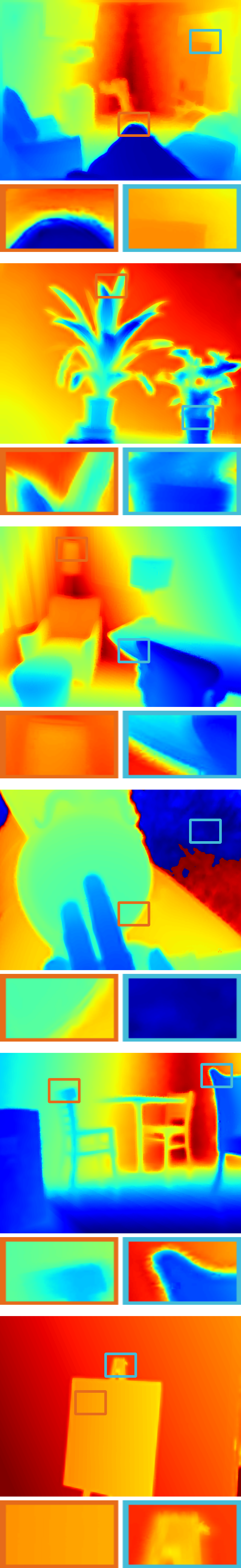}
    		\end{minipage}
    		\hspace{-0.08in}
    	}\subfigure[\scriptsize{DGF~\cite{DGF}}]{
    		\begin{minipage}[b]{0.12\linewidth}
  		 	\includegraphics[width=1\linewidth]{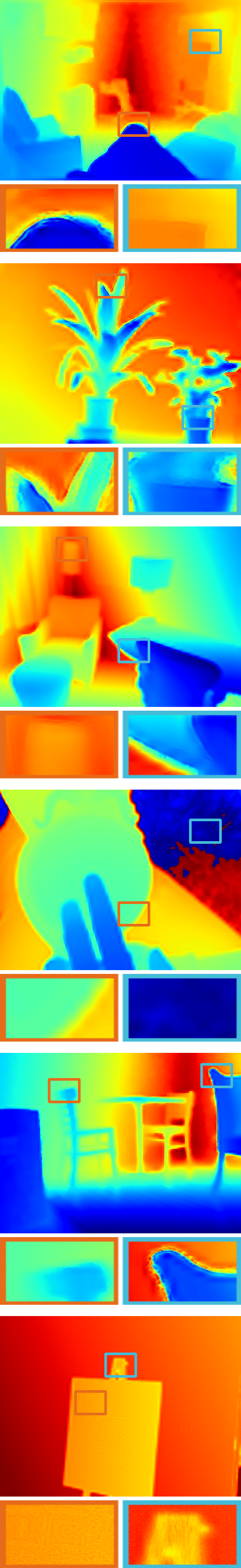}
    		\end{minipage}
    		\hspace{-0.08in}
    	}\subfigure[\scriptsize{GraphSR~\cite{GraphSR}}]{
    		\begin{minipage}[b]{0.12\linewidth}
  		 	\includegraphics[width=1\linewidth]{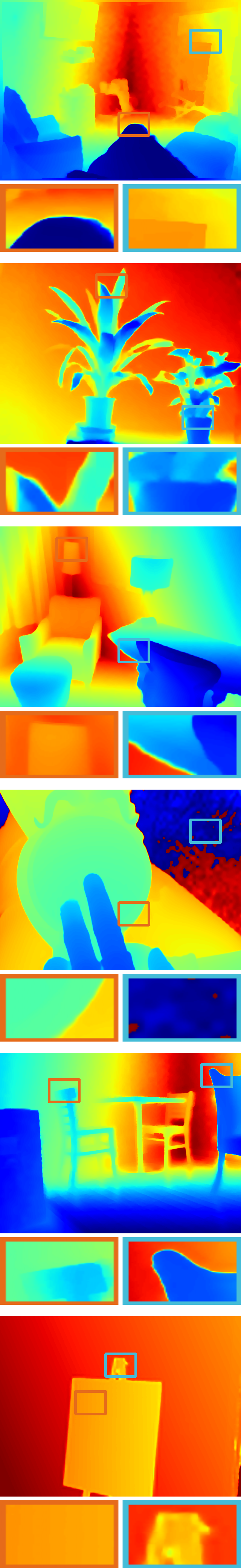}
    		\end{minipage}
    		\hspace{-0.08in}
    	}\subfigure[\scriptsize{AHMF~\cite{AHMF}}]{
    		\begin{minipage}[b]{0.12\linewidth}
  		 	\includegraphics[width=1\linewidth]{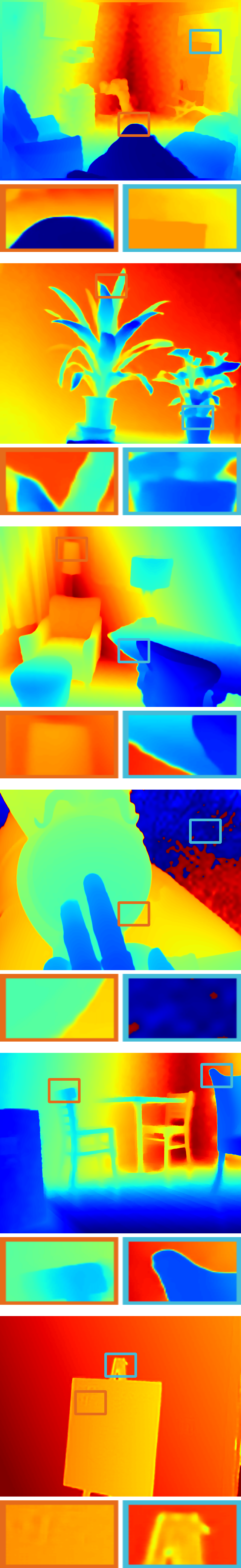}
    		\end{minipage}
    		\hspace{-0.08in}
    	}\subfigure[\scriptsize{JIIF~\cite{JIIF}}]{
    		\begin{minipage}[b]{0.12\linewidth}
  		 	\includegraphics[width=1\linewidth]{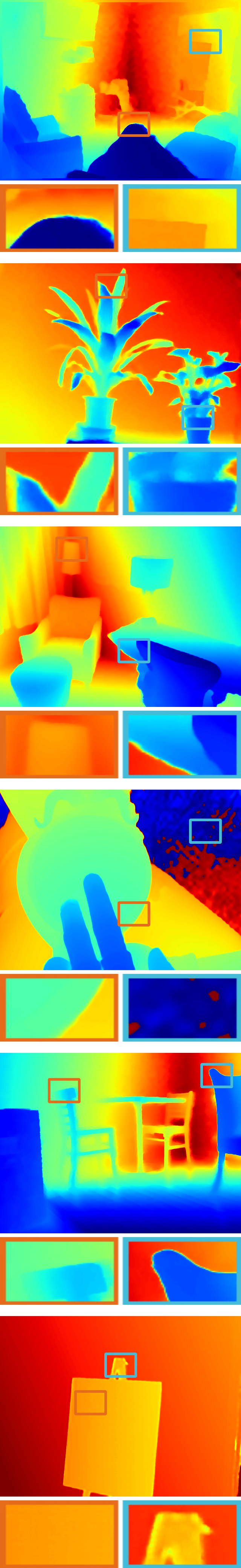}
    		\end{minipage}
    		\hspace{-0.08in}
    	}\subfigure[\scriptsize{GT}]{
    		\begin{minipage}[b]{0.12\linewidth}
  		 	\includegraphics[width=1\linewidth]{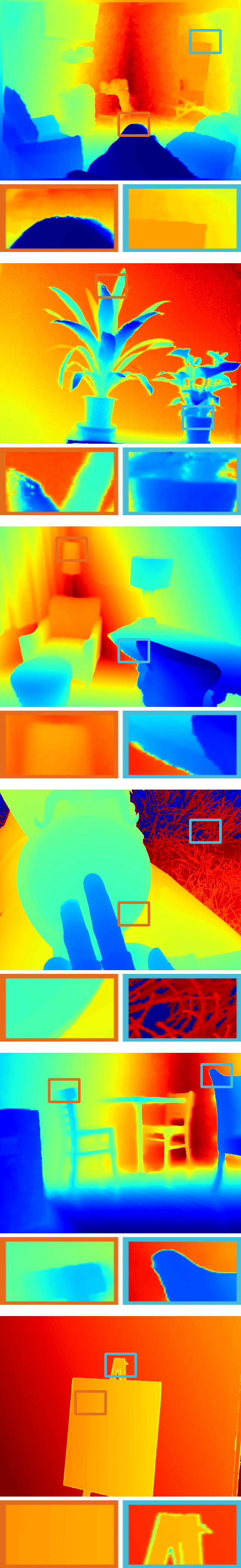}
    		\end{minipage}
    	}
	\end{center}
	\vspace{-.2in}
	\caption{Visual comparisons for $8\times$ guided depth map super-resolution. Top to bottom: Each row shows the super-resolved depth maps on the NYU v2~\cite{NYU}, RGB-D-D~\cite{FDSR}, SUN RGB-D~\cite{SUN}, MPI Sintel Depth~\cite{butler2012naturalistic}, DIML~\cite{cho2021deep} and DIDOE~\cite{DIDOE} datasets, respectively. Please enlarge the PDF for more details.}
	\label{fig:ecl}
	\vspace{-0.2in}
\end{figure}

For NYU v2~\cite{NYU} and RGB-D-D datasets~\cite{FDSR}, we mainly use them to compare the recently proposed learning-based methods, which include GF~\cite{GF}, DGF~\cite{DGF}, DMSG~\cite{DMSG}, DJF~\cite{DJF}, DJFR~\cite{DJFR}, DepthSR~\cite{DepthSR}, CUNet~\cite{CUNet}, SVLRM~\cite{SVLRM}, JIIF~\cite{JIIF}, FDKN~\cite{DKN}, DKN~\cite{DKN}, AHMF~\cite{AHMF}, FDSR~\cite{FDSR}, DCTNet~\cite{DCTNet} and GraphSR~\cite{GraphSR}. The RMSE values for different methods are presented in Tab.~\ref{md_rmse} and Tab.~\ref{tab:rd_rmse}. From these tables, we can see that the learning-based methods achieve significantly better performance than traditional GDSR models, which demonstrates the powerful learning ability of deep neural networks. In addition to quantitative comparisons, we also present the qualitative results of different methods in Fig.~\ref{fig:ecl}. As can be seen, the results of GF~\cite{GF} are over-smooth. To improve the performance of GF~\cite{GF}, \citet{DGF} propose to utilize two networks to modify the guidance and low-resolution depth images. The modified guidance and depth images are more suitable than the original ones. However, DGF~\cite{DGF} still cannot generate satisfactory results, since it still follows the basic framework of GF~\cite{GF} . For comparison, the deep learning based methods can produce results with sharper boundaries and less artifacts. From Fig.~\ref{fig:ecl} and Tab.~\ref{nyu_rmse}, we can also find that the RMSE values do not always consist with subjective assessment results. Thus, it is of vital importance to develop a more reasonable evaluation metric for the GDSR task. Finally, we report the parameter number, maximum GPU memory consumption, and average running times of different methods in Fig.~\ref{rmse_time}. As can be seen, models with more parameters can obtain better performance, but at the expense of high computational complexity. These models may not be suitable for real-world scenarios, especially for the resource-limited devices such as mobile phone. Therefore, it is also important to design light-weight and efficient models for the GDSR task. 

% http://192.168.113.35:8887/lab/tree/Jupyter/nyu_experiment/%E4%B8%BB%E8%A7%82%E6%95%88%E6%9E%9C.ipynb
\textbf{Bag of tricks for GDSR}. Recently, with the development of deep convolution neural networks, a lot of deep learning-based methods have been proposed and dominated the area of GDSR. By reviewing these methods, we find that most of them pay more attention to designing effective network architectures, and only a few works concern the training strategies which have been shown to be important for various tasks, such as image classification~\cite{he2019bag} and image super-resolution~\cite{lin2022revisiting}. To this end, we survey several tricks widely used in other vision tasks and explore whether they can improve the performance of GDSR. Specifically, we use the recently proposed AHMF~\cite{AHMF} as the baseline model and perform the following strategies one-by-one:
1) use geometric self-ensemble~\cite{lim2017enhanced} during inference; 2) employ mix-up as data augmentation method; 3) replace PReLU~\cite{he2015delving} with SiLU~\cite{hendrycks2016gaussian, elfwing2018sigmoid} activation; 4): during training, increase the patch size from $256 \times 256$ to $384 \times 384$ to enlarge the input \textit{field-of-view}; 5): augment the guidance image by randomly change brightness, contrast, saturation, and hue; 6) use cosine annealing learning rate decay to adjust the learning rate. The ablation studies of different training strategies are presented in Tab.~\ref{tab:abl}. We can observe that almost all the tricks contribute to the final performance, and by stacking all the tricks, the final model (+All of the above in Tab.~\ref{tab:abl}) can even outperform the current SOAT method JIIF~\cite{JIIF}.
\begin{table}[!t]\setlength{\tabcolsep}{2.9pt}\renewcommand{\arraystretch}{1.1}\renewcommand\theadfont{\footnotesize}
    \footnotesize 
    \centering
    \caption{Ablation studies of different training strategies on NYU v2 dataset~\cite{NYU} for $8\times$ depth map super-resolution. We use the AHMF~\cite{AHMF} as baseline. For the RMSE, the lower values means the better performance. The best performance is shown in \textbf{bold}, while the second and the third best performance are the \underline{underscored} and the \uwave{waved} ones, respectively.}
    \vspace{-0.1in}
    \begin{tabular}{lcccccccccccc}
		\toprule
        \multirow{2}{*}{Configuration} & \multicolumn{2}{c}{NYU v2} & \multicolumn{2}{c}{Sintel} & \multicolumn{2}{c}{DIDOE}  & \multicolumn{2}{c}{SUN RGB} & \multicolumn{2}{c}{RGB-D-D}& \multicolumn{2}{c}{DIML}\\
        \cmidrule(lr){2-3} \cmidrule(lr){4-5} \cmidrule(lr){6-7}  \cmidrule(lr){8-9}  \cmidrule(lr){10-11} \cmidrule(lr){12-13}
        ~ & RMSE & $\Delta$ & RMSE & $\Delta$ & RMSE & $\Delta$ & RMSE & $\Delta$ & RMSE & $\Delta$ & RMSE & $\Delta$\\
        \midrule
        Baseline & 2.887 & - & 5.624 & - & 6.143 & - & 0.985 & - &  1.733 & - & 1.699 & - \\
        +Self-ensemble & 2.827 & -0.060 & 5.577 & -0.027 & 6.088 & -0.055 & \uwave{0.954} & \uwave{-0.031} & \uwave{1.705} & \uwave{-0.028} & \uwave{1.677} & \uwave{-0.022}   \\
        +MixUp & 2.847 & -0.030 & \underline{5.592} & \underline{-0.032} & 6.185 &  +0.042 & \underline{0.945} & \underline{-0.040} & \textbf{1.695} & \textbf{-0.038} & \underline{1.672} & \underline{-0.027}  \\
        +SiLU Activation& 2.883 & -0.004 & 5.616 & -0.008 & \uwave{6.080} & \uwave{-0.063} & 0.984 & -0.001 & 1.741 & +0.008 & 1.685 & -0.014 \\
        +Large Patches & \uwave{2.825} & \uwave{-0.062} & 5.613 & -0.011 & \underline{6.059} & \underline{-0.084} & 0.981 & -0.004 & 1.756 & +0.023 & 1.683 & -0.016 \\
        +Color Augmentation & 2.892 & +0.005 & \uwave{5.594} & \uwave{-0.030} & 6.098 & -0.045 & 0.983 & -0.002 & 1.730 & -0.003 & 1.687 & -0.012 \\
        +Cosine Annealing & \underline{2.809} & \underline{-0.078} & 5.612  & -0.012 & 6.090 & -0.053 & 0.968 &  -0.017 & 1.728 & -0.005 & 1.685 & -0.014 \\
        +All of the above & \textbf{2.731} &\textbf{-0.156} & \textbf{5.558} & \textbf{-0.066} &\textbf{5.965} & \textbf{-0.178} & \textbf{0.916} & \textbf{-0.069} & \underline{1.701} & \underline{-0.032} & \textbf{1.628} & \textbf{-0.071} \\
		\bottomrule
		\end{tabular}
    \label{tab:abl}
    \vspace{-0.2in}
\end{table}

% \section{Other considerably related works}\label{other}

% \textit{In this section, we list some of the works that we encountered
% along the way that does not fit any category mentioned earlier.
% Some of them are not even tackling the SIDE problem as we
% defined. Nevertheless, they are considered interesting}

% \subsection{Single Depth Map Super-resolution}

% \subsection{Depth Completion}

% \subsection{Pan Shaprpening}

\section{Discussion and Conclusions}\label{conclusion}
In this paper, we present, to the best of our knowledge, the first comprehensive overview of guided depth map super-resolution (GDSR) methods. We first review the widely used datasets and evaluation metrics for GDSR. Then, we roughly classify existing GDSR methods into three categories: filtering-based methods, prior-based methods, and learning-based methods. In each category, we investigate representative and milestone methods and discuss their contributions, benefits, and weaknesses. Moreover, based on the unified experimental configurations, we provide a comprehensive evaluation of existing state-of-the-art GDSR methods.

It is true that the methods mentioned above have achieved gratifying performance and have significantly promoted the development of GDSR. However, we cannot neglect that there are still some challenging problems in GDSR. In the following, we list some of the challenges and present prospects for future research.

\textbf{Lightweight Models.} Some newly released smartphones are equipped with a range sensor to capture depth maps for face recognition, AR games, etc. Therefore, an accurate and efficient model is highly desirable to improve the quality of captured depth maps. However, current state-of-the-art GDSR methods are designed to be complicated to increase the model capacity for higher performance, making them hard to apply directly on the mobile device. To address this problem, model compression and neural architecture search can be adopted to learn a compact GDSR model with promising reconstruction accuracy.

\textbf{Evaluation Metrics}. Evaluation metrics play a fundamental role in computer vision tasks. First, the choice of objective functions is typically influenced by the evaluation metric. For instance, the $\mathcal{L}_1$ and $\mathcal{L}_2$ losses have been extensively used in GDSR as they are particularly relevant with the MAE and RMSE metrics. Then, newly proposed algorithms are needed to compare with existing methods using these metrics. However, previous studies~\cite{johnson2016perceptual, ledig2017photo} show that these metrics cannot accurately reflect the visual quality of the restored image. In addition, MAE and RMSE are full-reference image quality assessment scores, which cannot be used in real-world applications. Hence, it is an incredibly urgent and important task to develop appropriate evaluation metrics for GDSR.

\textbf{Scale Arbitrary GDSR.} Depth information has been widely used in a variety of tasks. Therefore, it is desirable to develop a GDSR model that can be adopted to upsample a given LR depth map to any scale. Current GDSR methods can only be applied to one or a limited number of integer upsampling factors. Recently, the implicit neural representation technique is employed in single image super-resolution~\cite{chen2021learning} for arbitrary scale super-resolution. The question of how to design a framework that incorporates the implicit neural representation technique with GDSR models may be a valuable topic for future research.

\textbf{Interact with High-level Tasks.} In most cases, the purpose of GDSR is not only to restore a high-resolution depth map with lower MAE values. Actually, we hope that the reconstructed depth map can facilitate the relevant high-level tasks, such as image segmentation and scene understanding. Therefore, we suggest employing the accuracy of high-level tasks as evaluation metrics to evaluate the performance of the GDSR method. Moreover, we can design a hybrid model which combines the GDSR model with other high-level tasks to make them learn from each other.

\textbf{Weakly-/Unsupervised Learning}. It is known to all that the deep learning-based approaches do not perform well in real-world cases. This is mainly because most of these methods require pairs of low- and high-resolution for training. However, the degradation process in the real-world scenario is so complex that it cannot be simulated by Bicubic or Nearest downsampling methods, which leads to a big gap between these synthetic depth maps and real-world low-resolution depth maps. In addition, it is time-consuming to develop a large-scale dataset which contains real LR and HR depth maps for GDSR models. Although some weak-/ un-supervised GDSR methods ara proposed,the performance of them is still worse than supervised methods. Therefore, employing the weakly-/unsupervised learning methods for GDSR appears to be a promising research direction.

\textbf{RGB-D Datasets.} Generally, the paired RGB-D data play an important role in the GDSR task, especially for the learning methods. In recent years, several RGB-D datasets have been released, greatly promoting the development of the GDSR algorithm. However, compared with the dataset for other computer vision tasks, such as object detection and classification, the size of these datasets is very small. In addition, most of the RGB-D datasets are captured from indoor scenes. Therefore, it is of vital importance to construct large-scale and realistic RGB-D datasets that can serve as benchmark datasets for future research.

% However, there has been
% little research on this task hence this needs be further investigated.

% Further improving semi-supervised or unsupervised methods to learn deblurring models appears to be a promising research direction.

% \begin{acks}
% \end{acks}


\section{Acknowledgements}
This work was supported by National Natural Science Foundation of China under Grants 92270116 and 62071155.

\normalem
\bibliographystyle{ACM-Reference-Format}
\bibliography{sample-simple-base}

\end{document}